\documentclass[journal,compsoc]{IEEEtran}

\usepackage{hyphenat}

\ifCLASSOPTIONcompsoc
  \usepackage[nocompress]{cite}
\else
  \usepackage{cite}
\fi
\usepackage[fleqn]{amsmath}

\usepackage{amssymb}

\ifCLASSINFOpdf
  \usepackage[pdftex]{graphicx}
\else
  \usepackage[dvips]{graphicx}
\fi

\usepackage{floatrow}
\newfloatcommand{capbtabbox}{table}[][\FBwidth]

\usepackage[monochrome]{color}
\usepackage{times}    % comment if you want LaTeX's defaulet font
\usepackage{url}      % llt: nicely formatted URLs
\usepackage{subcaption}
\usepackage{booktabs}

\usepackage{multicol}

\usepackage[pdftex]{hyperref}
\hypersetup{
pdftitle={Exploring the Generalizability of Spatio-Temporal Traffic Prediction: Meta-Modeling and an Analytic Framework},
pdfauthor={Leye Wang, Di Chai, Xuanzhe Liu, Liyue Chen, Kai Chen},
pdfkeywords={differential privacy, crowdsensing},
bookmarksnumbered,
pdfstartview={FitH},
colorlinks,
citecolor=black,
filecolor=black,
linkcolor=black,
urlcolor=black,
breaklinks=true,
}

\newcommand{\citep}[1]{{\cite{#1}}} 

\begin{document}

% \title{Differential Privacy for Sparse Mobile Crowd-Sensing with Even-Obfuscation Data Uncertainty-Minimization}

%\title{Bringing Differential Location Privacy to Sparse Crowdsensing with Consideration of Data Quality}

\title{Exploring the Generalizability of Spatio-Temporal Traffic Prediction: Meta-Modeling and an Analytic Framework}

\author{Leye~Wang,
        Di~Chai,
        Xuanzhe~Liu,
        Liyue~Chen,
        Kai~Chen
        % <-this % stops a space
\IEEEcompsocitemizethanks{\IEEEcompsocthanksitem L. Wang, X. Liu, and L. Chen are with Key Lab of High Confidence Software Technologies (Peking University), Ministry of Education, China, and Software Institute, Peking University, China.\protect\\
	E-mail: leyewang@pku.edu.cn, liuxuanzhe@pku.edu.cn, chenliyue2019@gmail.com
	\IEEEcompsocthanksitem D. Chai and K. Chen are with Hong Kong University of Science and Technology, Hong Kong SAR, China.\protect\\
	E-mail: dchai@connect.ust.hk, kaichen@cse.ust.hk
}% <-this % stops an unwanted space
}

% \numberofauthors{3}
% \author{
%   \alignauthor 1st Author Name\\
%     \affaddr{Affiliation}\\
%     \affaddr{Address}\\
%     \email{e-mail address}\\
%     \affaddr{Optional phone number}
%   \alignauthor 2nd Author Name\\
%     \affaddr{Affiliation}\\
%     \affaddr{Address}\\
%     \email{e-mail address}\\
%     \affaddr{Optional phone number}    
%   \alignauthor 3rd Author Name\\
%     \affaddr{Affiliation}\\
%     \affaddr{Address}\\
%     \email{e-mail address}\\
%     \affaddr{Optional phone number}
% }
%\markboth{IEEE TKDE}
{}

\IEEEcompsoctitleabstractindextext{%
\begin{abstract}
The Spatio-Temporal Traffic Prediction (STTP) problem is a classical problem with plenty of prior research efforts that benefit from traditional statistical learning and recent deep learning approaches. While STTP can refer to many real-world problems, most existing studies focus on quite specific applications, such as the prediction of taxi demand, ridesharing order, traffic speed, and so on. This hinders the STTP research as the approaches designed for different applications are hardly comparable, and thus how an application-driven approach can be generalized to other scenarios is unclear. To fill in this gap, this paper makes three efforts: (i) we propose an analytic framework, called STAnalytic, to qualitatively investigate STTP approaches regarding their design considerations on various spatial and temporal factors, aiming to make different application-driven approaches comparable; (ii) we design a spatio-temporal meta-model, called STMeta, which can flexibly integrate generalizable temporal and spatial knowledge identified by STAnalytic, (iii) we build an STTP benchmark platform including ten real-life datasets with five scenarios to quantitatively measure the generalizability of STTP approaches. In particular, we implement STMeta with different deep learning techniques, and STMeta demonstrates better generalizability than state-of-the-art approaches by achieving lower prediction error on average across all the datasets.
\end{abstract}
\begin{keywords}
spatio-temporal prediction; crowd flow; meta-model
\end{keywords}
}

\maketitle

\IEEEdisplaynotcompsoctitleabstractindextext
\IEEEpeerreviewmaketitle

\renewcommand{\arraystretch}{1.3}

%!TEX root = stmeta.tex

\section{Introduction}

The \textbf{Spatio-Temporal Traffic Prediction (STTP)} problem refers to the  problem that resides in many urban predictive applications related to both spatial and temporal human mobility dynamics, e.g., the predictions of taxi demand \citep{zhang2017deep}, metro human flow \citep{wei2012forecasting}, electrical vehicle charging usage \citep{bae2011spatial}, and traffic speed \citep{li2017diffusion}. The STTP problems play a vital role in today's smart city management and organization, e.g., traffic monitoring and emergency response.

Traditionally, STTP is often formulated as a time series prediction problem, where statistical methods such as ARIMA (AutoRegressive Integrated Moving Average) \citep{williams1998urban,hamed1995short} are widely used. In recent years, with the advance of machine learning, especially deep learning techniques, a variety of new models have been developed for STTP. Today, researchers can be quickly overwhelmed by numerous STTP related papers that continuously emerge in top-tier conferences and journals \citep{zhang2017deep,li2017diffusion,liang2018geoman,geng2019spatiotemporal,chai2018bike}. However, most efforts focus on the sophisticated application-specific model design and then test their models on limited data of one or few specialized applications, e.g., ridesharing~\citep{geng2019spatiotemporal}, bikesharing~\citep{chai2018bike,yang2016mobility}, and highway traffic speed \citep{li2017diffusion}. Although the proposed models can conceptually be applied to other STTP scenarios, whether the performance can still be as good as their specialized applications remains unclear. In other words, it becomes rather difficult to analyze whether an STTP model can be quickly generalized over various scenarios. Indeed, the generalizability is a fundamental and key property of an STTP model, and significantly determines the potential impact scope. In this regard, it is highly urgent to require a general analytic framework that can help investigate and compare different application-driven STTP models. Furthermore, the insights derived from the analytic framework can guide researchers to justify the generalizability of these models, and then design new models with better generalizability.

To fill this research gap, this paper makes efforts from the following aspects:

1) To make different application-driven STTP approaches comparable, we propose \textbf{an analytic framework}, called \textit{STAnalytic},  to investigate the STTP approaches from their considered high-level spatial and temporal factors. Particularly, STAnalytic maps an STTP model into a two-level hierarchical analysis process where the first level is \textit{spatial and temporal} perspective, and the second level is \textit{knowledge and modeling} perspective. Then, every STTP model can be analyzed as two research questions: (1) \textit{``what effort is done from temporal and spatial perspective?''}; and (2) \textit{``which spatial/temporal knowledge is considered with which modeling techniques?''}. This qualitative analysis can provide useful insights about whether an STTP model can be well generalized to various scenarios even before we run quantitative experiments over the model.

2) To elaborate the effectiveness of STAnalytic in terms of helping design generalizable STTP models, we propose a \textbf{spatio-temporal meta-model}, called \textit{STMeta}, based on the analysis results of the STTP over state-of-the-art in literature \citep{zhang2017deep,li2017diffusion,liang2018geoman,geng2019spatiotemporal}. STMeta is a ``model of model'' (meta-model) which has a hierarchical structure to flexibly and efficiently integrate generalizable temporal and spatial knowledge identified by STAnalytic from literature. With state-of-the-art deep learning techniques such as graph convolution \citep{defferrard2016convolutional} and attention mechanisms \citep{vaswani2017attention}, we implement three variants of STMeta and verify their generalizable effectiveness via multiple real-life  STTP benchmark datasets (listed next). %The experiments with our benchmark datasets verify that the STMeta variants are more generalizable than  state-of-the-art STTP approaches designed for specific scenarios. %Across all the scenarios, the best STMeta variant can reduce on average 5\% of prediction error compared to the best existing approach \citep{geng2019spatiotemporal}.

3) To alleviate the issue that today's STTP research studies usually conduct experiments on certain specific applications and do not justify the generalizability among various applications quantitatively, we build \color{blue} a set of \textbf{STTP benchmark datasets} \color{black} including five scenarios, i.e., ridesharing, bikesharing, metro, electric vehicle charging, and traffic speed. The datasets cover ten cities, where the longest time duration spans four years and the largest number of traffic data records is more than 400 million. We have released our code and data repository\footnote{\url{https://github.com/uctb/UCTB}}.%\footnote{Please visit the code and data repository at \url{https://uctb.github.io/UCTB/}.}

To summarize, this paper makes the following contributions:
  
  1) We develop an analytic framework, namely \textit{STAnalytic}, following which existing STTP approaches can be qualitatively explored. To the best of our knowledge, this is one of the first studies that propose such an analytic framework for the STTP problem. 

  2) We design a meta-model called \textit{STMeta} based on the insights derived from STAnalytic over state-of-the-art research \citep{zhang2017deep,li2017diffusion,liang2018geoman,geng2019spatiotemporal}. STMeta can flexibly and efficiently take into account multiple spatial and temporal knowledge for building an STTP model. %Thanks to the feasibility of STMeta to integrate various spatial and temporal knowledge, STMeta can be further leveraged to test the generalizability of different knowledge factors in literature.
  
  %Note that STMeta is a meta model and we can implement it with various state-of-the-art deep learning techniques such as graph convolution \citep{defferrard2016convolutional,li2017diffusion} and attention mechanisms \citep{velivckovic2017graph,vaswani2017attention}.

  3) We build a set of real-life STTP benchmark datasets covering five scenarios and ten cities. The benchmark results verify that STMeta generally outperforms state-of-the-art approaches. After analyzing the results, we provide several design guidelines to develop STTP approaches with better generalizability.

%!TEX root = stmeta.tex

\section{Problem and Generalizability}

\subsection{Problem Formulation}

First, we formulate the \textbf{Spatio-Temporal Traffic Prediction} problem.  Suppose that there are a set of $n$ locations $\mathcal L = \{l_1, l_2, \cdots, l_n\}$, and for each $l_i \in \mathcal L$, it has a historical series of traffic information from time slot $1$ to the current slot $k$, denoted as  $F_i = \{f_{i1}, f_{i2},\cdots, f_{ik}\}$. Then, we want to predict the traffic information for each $l_i$ in the next time slot $k+1$ to minimize,
\begin{equation}
	error(\hat f_{i(k+1)}, f_{i(k+1)}) \quad \forall l_i \in \mathcal L
\end{equation}
where $\hat f_{i(k+1)}$ is the predicted traffic of $l_i$ in the next time slot $k+1$, and $f_{i(k+1)}$ is the ground truth; the $error$ function may have many choices. In this paper, we use RMSE (root mean square error).

\subsection{Application Scenarios}

STTP is an abstraction of many real-world  smart city applications that are reported and studied in the literature:
\begin{itemize}

\item \textit{Region-based Ridesharing Demand Prediction} \cite{geng2019spatiotemporal}: With the popularity of ridesharing services, one fundamental problem is to predict how many ridesharing demands will emerge in every area of a city. Usually, a city will be split into a set of equal-size regular regions (e.g., grids of 1km * 1km) \cite{geng2019spatiotemporal,zhang2017deep} or irregular regions (e.g., functional areas split by road network) \cite{sun2020predicting}, and then one needs to predict the ridesharing demand for each region in near future. %Taking each grid as a location, then the grid-based ridesharing demand prediction is an STTP problem.

\item \textit{Station-based Bikesharing Demand Prediction} \citep{chai2018bike}: Bikesharing is another quickly developed service in many cities nowadays. One major type of bikesharing services builds a set of fixed stations in a city and then users can borrow and return bikes at any of those stations. As each station has a limited number of docks to hold bikes, it is important to predict how many users will borrow or return bikes in near future.%, so that the bikesharing company can re-balance the distributions of bikes in advance (e.g., by truck). %Taking each bike station as a location, the station-based bikesharing demand prediction is an STTP problem.

\item \textit{Other Applications.} In literature, a lot of other specific problems can also be converted to STTP problems. Actually, most of them can be categorized to region-based or station-based problems. For example, for taxi flow prediction of a city, the problem is often formulated same as ridesharing demand and the regions are pre-defined \cite{zhang2017deep,Zhang0QLYL18}; for traffic speed prediction of road segments, the problem formulation is similar to the station-based bikesharing case~\citep{yu2018spatio,Zheng2020GMANAG}.%,Wu2019GraphWF}.
 
\end{itemize}

\subsection{Generalizability}

While most prior studies focus on specific applications, this paper aims to explore the generalizability issue in STTP. In particular, we concentrate on two generalizability issues: \textit{model generalizability} and \textit{knowledge generalizability}:
%First, considering various applications, we aim to design a \textit{generalizable} STTP model:

\textbf{Model Generalizability}. As different STTP applications involve diverse spatial and temporal knowledge factors, the model generalizability refers to: (1) \textit{Regarding different spatial and temporal knowledge factors,  is a model generalizable to incorporate diverse factors?} (2) \textit{By incorporating different factors, can a model be generally competitive to the state-of-the-art models in a variety of applications?}

%our proposed model expects to have a flexible structure to encode different factors in various literature applications. Furthermore, by incorporating various factors, our model also aims to achieve the competitive accuracy compared to the state-of-the-art approaches in different applications.

%Second, with the generalizable model, we can investigate how different spatial and temporal knowledge factors are \textit{generalizable} among different applications.

\textbf{Knowledge Generalizability}. With the generalizable model that can incorporate various spatial and temporal knowledge factors, we aim to study the knowledge generalizability issue: \textit{Given a certain spatial or temporal knowledge factor, is it generalizable to be effective for various STTP applications?}

To address the \textit{model generalizability} issue, we first propose \textit{STAnalytic}, an analytic framework to investigate the spatial and temporal factors considered in a specific STTP model, which can facilitate a qualitative analysis of the model's generalizability. With the spatial and temporal factors identified by STAnalytic in mind, we propose \textit{STMeta}, a meta-model that can incorporate distinct factors in a unified manner; we verify STMeta's superiority of generalizability over state-of-the-art models by running \textit{an extensive benchmark experiment} with ten real-life STTP datasets. Finally, with STMeta, we tackle the \textit{knowledge generalizability} issue by studying how the importance of different spatial and temporal factors vary with the change of benchmark datasets.
\color{black}
%!TEX root = stmeta.tex

\begin{figure}[t]
	\centering
	\includegraphics[width=1\linewidth]{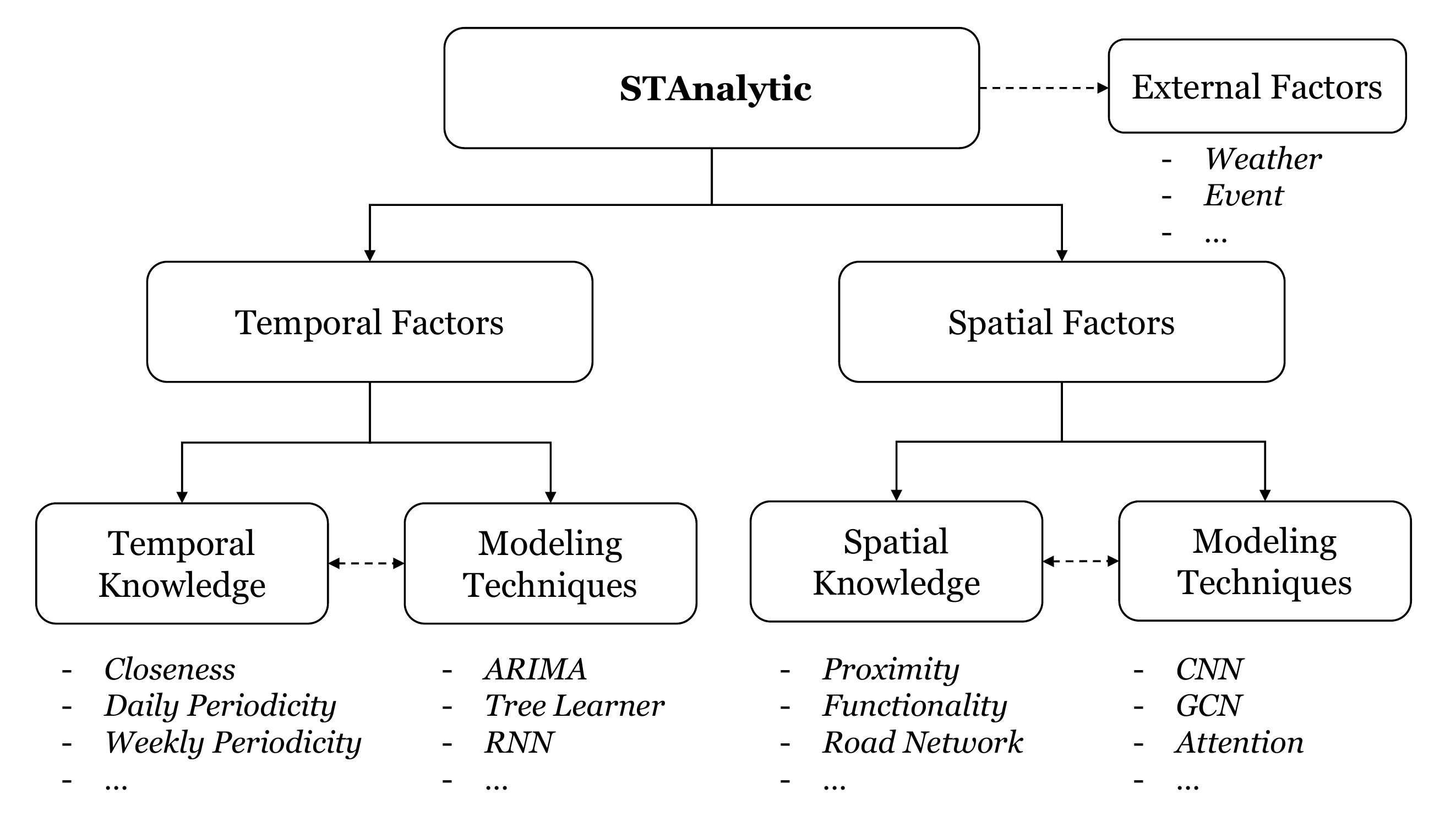}
	\caption{STAnalytic: an analytic framework for STTP.}
	\label{fig:framework}
	\vspace{-1em}
\end{figure}

\section{STAnalytic: Analytic Framework}% for Traffic Prediction}
%As the STTP problem can be formulated as a high-level abstraction by capturing the commonality, it is ideal that a model designed for one application can be quickly generalized to another. However, the preceding mentioned efforts provide effective solutions for very specific contexts and scenarios, and the generalizability is hard to justify.

 We propose an analytic framework, called \textit{STAnalytic}, with which we can investigate and compare different application-driven STTP models and justify their generalizability.

\subsection{Framework Overview}

Our framework \textit{STAnalytic} is illustrated in Figure~\ref{fig:framework}. Overall, the framework considers STTP approaches mainly from two aspects: \textit{temporal} and \textit{spatial} factors by answering the following research questions.

\textbf{RQ1}: \textit{Does the approach consider temporal and/or spatial factors in predicting traffic?}

\textbf{RQ2}: \textit{What temporal knowledge factors have been taken into building the approach? For each considered temporal knowledge factor, how does the approach model it?}

\textbf{RQ3}: \textit{What spatial knowledge factors have been taken into building the approach? For each considered spatial knowledge factor, how does the approach model it?}

Regarding RQ1, the STTP approaches based on traditional time-series analysis techniques, such as autoregressive integrated moving average (ARIMA), consider only temporal factors \citep{hamilton1994time,hamed1995short}. Other studies leveraging more advanced temporal learning techniques, such as long short-term memory (LSTM) \cite{ma2015long,tian2015predicting}, also belong to this temporal-factor-only category. In comparison, more recent STTP approaches mostly consider both temporal and spatial factors explicitly into the model design, especially with state-of-the-art deep learning techniques such as convolution networks \citep{zhang2017deep,chai2018bike}.

Regarding RQ2 and RQ3, it needs more effort to answer for an STTP approach. To help answer RQ2 and RQ3, the next subsection will review and summarize common temporal and spatial factors that have been considered in the STTP literature. Then, when analyzing a new STTP approach, we can quickly check `\textit{whether the temporal and spatial factors in consideration belong to our summarized ones}' and `\textit{whether there is new temporal or spatial knowledge beyond our summarized ones}', so as to answer RQ2 and RQ3.

While spatial and temporal factors are the foundation for traffic prediction, other \textit{external factors}, e.g., weather \cite{koesdwiady2016improving} and social events \cite{ni2016forecasting}, may impact traffic patterns prominently, leading to the fourth research question.

\textbf{RQ4:} \textit{What external factors have been considered in predicting traffic?}

\subsection{Temporal and Spatial Knowledge Factors}
\label{sub:st_factors}

To help answer RQ2 and RQ3, we investigate the temporal and spatial knowledge that has been well studied in the literature.

\subsubsection{Temporal Knowledge Factors and Modeling} 

Modeling temporal factors has been well studied in STTP, which can be traced back to a very classical research area, \textit{time-series analysis}~\citep{hamilton1994time}. In general, most time series analysis techniques can be applied to STTP; for clarity, we just highlight some most widely used temporal knowledge and its related techniques. The most widely considered temporal knowledge includes temporal closeness, daily periodicity, and weekly periodicity:

\textbf{Temporal Closeness}. One of the most intuitive ways to predict future traffic data is checking back the traffic information of the recent time slots, which is called \textit{closeness}. In other words, for the models considering closeness (in fact, almost all the models consider closeness), they would take the recent a few $k$ time slots' traffic data as input and then predicts the future. 

In literature, many classical statistical methods have been proposed to extract meaningful closeness patterns from the time-series data. The most famous model is perhaps ARIMA \citep{hamed1995short}. In brief, ARIMA models the traffic of a future slot as a linear combination of the flow of recent slots. As non-linear relations are hard to find with ARIMA, recently researchers started using non-linear models such as tree-based models (e.g., XGBoost~\cite{chen2016xgboost}) and recurrent neural networks (e.g., LSTM~\cite{gers1999learning,ma2015long}). Note that, regardless of the modeling techniques, if an approach takes recent a few time slots' traffic data as input, then its temporal consideration factor is closeness.

\textbf{Daily Periodicity}. As human activity has high regularity, the traffic dynamics, whatever the specific application, also often has obvious patterns. The daily periodicity is one of the most significant patterns. For example, on workdays around 8:00 a.m., people get out of home and go to workplaces, leading to the morning rush hours when many types of traffic data (metro, taxis, buses, etc.) will go from the residential area to the working area. In general, if a model considers daily periodicity, it will put the traffic data of previous days at the same time (e.g., the same hour)  into the input.

To model the daily periodicity, the \textit{seasonal} component can be added to ARIMA \citep{williams1998urban}. Another widely used method is selecting the historical traffic data at the same daily time slot during the last few days into the input, and then use non-linear models to extract daily patterns  \citep{zhang2017deep}.

\textbf{Weekly Periodicity}. Similar to daily periodicity, the traffic data of certain scenarios may follow weekly periodicity. As an example, Saturday's traffic will usually be much more similar to last Saturday instead of Friday.

In practice, the difference of modeling weekly periodicity compared to daily periodicity is the temporal lags. For example, suppose we want to predict the traffic data of `\textit{8:00-8:30 a.m. Oct. 15}'. By considering daily periodicity, we take the traffic data of `\textit{8:00-8:30 a.m. Oct. 14}' as input; in comparison, by considering weekly periodicity, we take that of `\textit{8:00-8:30 a.m. Oct. 8}' as input. Apart from the different-lag inputs, the modeling techniques are almost the same for weekly and daily periodicity, such as seasonal ARIMA \citep{williams2003modeling}.

\subsubsection{Spatial Knowledge Factors and Modeling}

Compared to temporal knowledge modeling, spatial knowledge modeling recently attracts more interest from researchers as it is more complicated and heterogeneous. For different scenarios, the spatial knowledge may also be different from each other.

\textbf{Geographic Proximity}. `\textit{Everything is related to everything else. But near things are more related than distant things.}', pointed by Waldo R. Tobler in 1969, has been extensively recognized as the `First Law of Geography'. It clearly highlights the importance of proximity in geographic studies. To model geographic proximity in STTP, one widely used method is: for a target location, firstly select the time-series traffic data in near locations into the input, and then leverage methods such as K nearest neighbors to aggregate near locations' knowledge. More recently, convolution neural networks (CNN) have been widely used to extract geographic proximity patterns, especially for grid-based traffic prediction problems \cite{zhang2017deep}. CNN is good at this because it learns a small $m \times n$ parameter matrix (e.g., $3\times3$ or $5\times5$) that can well aggregate a (grid) location's data together with its nearby (grid) locations. As CNN can be stacked to a very deep structure (e.g., by ResNet~\cite{zhang2017deep}), then the geographic proximity pattern can be extracted at different levels.

\begin{table*}[t]
	\scriptsize
	\caption{Analysis of representative STTP approaches with STAnalytic.}
	\label{tab:existing_work}
	\begin{tabular}{@{}llllll@{}}
		\toprule
		& \textbf{Model Name} & \textbf{RQ2: Temporal} & \textbf{RQ3: Spaital} & \textbf{RQ4: External} & \textbf{Modeling Technique (T: Temporal, S: Spatial)} \\ \midrule
		\textbf{Temporal Only} &&&& \\
		\textit{Hamed et al. 1995} \cite{hamed1995short} & --- & close. & --- & none & ARIMA (T) \\
		\textit{Williams et al. 1998} \cite{williams1998urban} & --- & close., daily & --- &  none & Seasonal ARIMA (T) \\
		\textit{Williams et al. 2003} \cite{williams2003modeling} & --- & close., weekly & --- &  none & Seasonal ARIMA (T) \\
		\textit{ Ma et al. 2015} \cite{ma2015long} & --- & close. & ---&  none & LSTM (T) \\ \midrule
		\textbf{Temporal \& Spatial} &&&& \\
		\textit{Zhang et al. 2017} \cite{zhang2017deep} & ST-ResNet & close., daily, weekly & prox. &  weather, holiday& Residual Convolution (T, S) \\
		\textit{Liang et al. 2018} \cite{liang2018geoman} &  GeoMAN & close. & func. & none & LSTM (T), Attention (S) \\
		\textit{Li et al. 2018} \cite{li2017diffusion} & DCRNN & close. & prox. &  none& GRU (T), Diffusion Convolution (S) \\
		 \textit{Yu et al. 2018} \cite{yu2018spatio} &  STGCN & close. & prox. &  none& Gated Convolution (T), Graph Convolution (S) \\		
		\textit{Geng et al. 2019} \cite{geng2019spatiotemporal} & ST-MGCN & close., daily, weekly & prox., func., conn. &  none & GRU (T), Graph Convolution (S) \\
		 \textit{Guo et al. 2019} \cite{guo2019attention} &  ASTGCN & close., daily, weekly & prox. &  none & Attention \& Graph Convolution (T, S) \\
		 \textit{Wu et al. 2019} \cite{wu2019graph} &  Graph-WaveNet & close. & prox., data-driven &  none & Gated Unit \& Dilated Convolution (T), Graph Convolution (S) \\
		 \textit{Bai et al. 2020} \cite{bai2020adaptive} &  AGCRN &  close. &   data-driven &  none &  GRU (T), Graph Convolution (S) \\
		 \textit{Zheng et al. 2020} \cite{zheng2020gman} &  GMAN &  close. &   prox. &  none &  Attention (T, S) \\
		 \textit{Song et al. 2020} \cite{song2020spatial} &  STSGCN &  close. &   prox. &  none &  Graph Convolution (T, S) \\  \textit{Yi and Park 2020}
		\cite{yi2020hypergraph} &  HGC-RNN &  close. &   func. &  holiday &  GRU (T), Hypergraph Convolution (S) \\
		
 \bottomrule
	\end{tabular}
\end{table*}

\textbf{Location Functionality}. Functionality is one fundamental character of a location. For example, some locations are residence areas, some locations are shopping areas, and others are industrial areas. Apparently, these characteristics, if obtained, can greatly improve our understanding of these locations' traffic patterns \citep{yuan2014discovering}.

In practice, one of the most widely used data sources to characterize the functionality of a location is points-of-interests (POIs) \citep{geng2019spatiotemporal} and/or their related social media open check-ins \citep{yang2013fine,wang2019cross}. For example, the number or the distribution of different types of POIs are often used as a spatial feature vector for a location in literature~\citep{guo2018citytransfer,liang2018geoman}. With such feature vectors, different modeling methods can be leveraged to extract the hidden patterns about the location functionality that are related to the traffic data \citep{yao2018deep}. Note that, if we have a large amount of historical data among many locations, it is probable that we can directly infer the location functionality from its historical traffic records, so some existing studies also develop methods to use traffic pattern or similarity to describe location functionality \citep{wang2019cross,Chen2017FineGrainedUE,Fan2014CitySpectrumAN}.

\textbf{Inter-Location Relationship}. In real-world  applications, there are many types of spatial relationships between different locations which may indicate traffic patterns. For example, in traffic volume prediction, the locations connected by the same major city road (e.g., circle road in Beijing) will probably have related flow patterns, i.e., \textit{connectivity} relationship \cite{geng2019spatiotemporal}; in metro station flow prediction, the stations in the same metro line may also hold certain correlations in the flow patterns, i.e., \textit{same-line} relationship.

To model such diverse inter-location relationships, a natural way is to build a graph to link locations with certain relationships. That is, we see each location as a graph node, and then link two location nodes with an edge if they have some relationship (e.g., connected by the same road). Recently, graph convolution techniques have become very powerful tools to extract hidden spatial knowledge from such constructed inter-location relation graphs~\cite{geng2019spatiotemporal,chai2018bike}. 

It is worth noting that, the \textit{geographic proximity} and \textit{location functionality} can also be seen as special instances of inter-location relationships. For geographic proximity, nearby locations can be linked in the graph; for location functionality, the locations with similar functionality can also be connected \cite{geng2019spatiotemporal}. \textbf{This indicates that the inter-location relationship is promising to be generalized to represent various spatial knowledge for STTP. }

\textbf{Data-Driven Modeling without Explicit Knowledge}. While spatial patterns of traffic are complicated and heterogeneous in reality, some recent studies attempt to catch spatial knowledge directly by data-driven methods rather than encoding explicit knowledge factors. One representative method is first randomly constructing an inter-location graph, and then applying the graph neural network learning technique to refine the inter-location graph \cite{wu2019graph,bai2020adaptive}. Generally, the patterns learned by such purely data-driven methods are highly dependent on the amount and quality of the input historical data. Besides, since the inter-location relationship becomes learnable, the overall computation overhead is increased \cite{bai2020adaptive}.

\subsection{External Factors}

Besides spatial and temporal factors, researchers have also studied how various external factors may influencer traffic patterns \cite{zheng2019deepstd}. These research achievements inspire that for a fine-grained traffic prediction, external factors could be a useful extra knowledge source. We list some representative external factors widely studied in the literature.

\textbf{Weather}. Weather information, including temperature, humidity, condition (e.g., rainy, cloudy), etc., would impact human mobility obviously. It has been reported that more than 20\% of car accidents are related to weather conditions.\footnote{\url{https://ops.fhwa.dot.gov/weather/q1_roadimpact.htm}} Some researchers have analyzed the correlations of various weather variables and traffic for identifying influential weather factors \cite{koesdwiady2016improving}.

\textbf{Workday/Holiday}. Human mobility patterns are  diversified between workdays and holidays. For instance, during the Thanksgiving and Christmas holiday, there would be a significant increase on long-distance travels in U.S., leading to heavy congestion on highways \cite{jun2010understanding}. Hence, whether it is a workday or holiday can cause distinct traffic flows.

\textbf{Event}. When certain events happen around a location, the corresponding traffic would be biased from its normal pattern. For instance, when a sports game or concert is held at a stadium, then the passenger flow around the stadium would rapidly rise up, thus impacting the traffic of its nearby metro \cite{ni2016forecasting}, bikesharing stations \cite{chen2016dynamic}, etc.

%A traditional statistical method to automatically catch spatial knowledge is vector autoregression (VAR) and its variants, which can catch the linear interdependencies between different time series data from various locations~\citep{chandra2009predictions}. More recently, with the deep learning techniques rapidly evolving, the attention mechanism also becomes one of the powerful modeling techniques to automatically find the spatial knowledge hidden in the time series data of different locations~\citep{liang2018geoman}. Generally, the patterns learned by such purely data-driven methods are highly dependent on the number and quality of the input historical data, and thus how to avoid overfitting to the training data is very important for the performance.

\color{black}

\subsection{Analyzing Existing Models}
\label{sub:analyze_existing}

With the preceding issues, now we summarize representative research studies on STTP with STAnalytic. Since in recent years we have witnessed numerous STTP efforts, %we cannot enumerate all of them due to the page limit. 
we thus only select some representative studies from both classical statistical learning and new deep learning methodologies to give an overview. %More specifically, these studies are selected that they have tried to introduce diverse temporal and/or spatial knowledge into their constructed approaches. 

Regarding RQ1, we select two main streams of traffic prediction studies to analyze with STAnalytic. The first stream only considers the temporal factors with classical statistical methods (e.g., ARIMA) or recent deep learning methods (e.g., LSTM); the second stream leverages the deep learning techniques with both temporal and spatial knowledge in consideration. 

As shown in Table~\ref{tab:existing_work}, most of the first stream of works are much older, including Hamed et al. (1995) \cite{hamed1995short}, Williams et al. (1998) \cite{williams1998urban}, and Williams et al. (2003) \cite{williams2003modeling}. All of them take the traffic prediction problem as time-series data analysis and thus solve the problem with ARIMA methods. The results from Williams et al. (1998, 2003) \cite{williams1998urban,williams2003modeling} reveal that the daily and weekly periodicity widely exists and thus it is important to consider them in the temporal factors of the traffic prediction.  More recently, researchers start leveraging advanced machine learning techniques, such as LSTM, to learn the temporal patterns in traffic prediction \cite{ma2015long}. \color{black}

 For the second stream of works, we list several representatives \cite{zhang2017deep,liang2018geoman,li2017diffusion,yu2018spatio,geng2019spatiotemporal,guo2019attention,wu2019graph,bai2020adaptive}, which are published in prestigious venues and highly cited. %\footnote{ Except certain papers published recently in 2020, all the other representative works have been cited for more than 200 times according to Google Scholar when we write the paper (Oct. 2021). Hence, these works can be seen as high-impact ones in the traffic prediction literature.}  
\color{black} All of these studies apply deep learning techniques into STTP with both temporal and spatial knowledge in consideration. Note that their modeling techniques are distinct (see `modeling technique' in Table~\ref{tab:existing_work}), thus not directly comparable from the technique perspective. However, with STAnalytic, we put more focus on analyzing which types of temporal and spatial knowledge are taken into consideration, and then these studies are clearly comparable as follows:

1) From temporal knowledge (RQ2), ST-ResNet \cite{zhang2017deep}, ST-MGCN \cite{geng2019spatiotemporal}, and  ASTGCN \cite{guo2019attention} \color{black} consider both closeness and daily/weekly periodicity, while others only consider closeness.

2) From spatial knowledge (RQ3), ST-MGCN \cite{geng2019spatiotemporal} considers multiple factors including proximity, functionality, and road connectivity. In comparison, other studies only consider one spatial knowledge of proximity or functionality.  Note that Graph-WaveNet \cite{wu2019graph} and AGCRN \cite{bai2020adaptive} use data-driven methods to automatically extract spatial knowledge, which has the potential to outperform specified spatial knowledge with adequate high-quality historical data.

\color{black}
Based on the above comparison, among these approaches investigated by STAnalytic, we can infer that \textit{ST-MGCN \cite{geng2019spatiotemporal} is probably more generalizable over various scenarios, as it considers more temporal and spatial factors which have been verified effective in literature}. That is, \textit{the more knowledge, the better prediction}. Later in Section~\ref{sec:evaluation}, we will quantitatively compare most of these representative approaches with real-life STTP datasets.%\footnote{GeoMAN \cite{liang2018geoman} learns one distinct deep model for each location, which is much more time-consuming than the other approaches learning one model for all the locations, and hard to be scaled up in practice when there can be hundreds/thousands of locations. So we do not include it in our experiments.}

Another interesting observation from the analysis is based on RQ4. While the literature has shown that various external factors, such as weather and events, can influence traffic patterns, most spatio-temporal traffic prediction methods ignore external factors completely. There may be several reasons. First, how to extract useful external factors, such as events, is a non-trivial problem itself \cite{ni2016forecasting}. Second, how external factors impact traffic patterns is highly application-dependent (e.g., the event impacts on metro passenger volume may not be generalized to traffic speed). Then, it is still challenging for general-purpose traffic prediction approaches to incorporate external factors in a unified way for diverse applications. According to this state and making the method comparison fair, we then ignore external factors for all the approaches in our quantitative benchmark experiment (Section~\ref{sec:evaluation}). Nevertheless, we believe that investigating external factors is a promising research direction for STTP.

\color{black}

%!TEX root = stmeta.tex

\section{STMeta: STTP Meta-Model}% with Multiple Spatial and Temporal Knowledge}

To show the effect of STAnalytic in guiding the design of generalizable STTP approaches, we propose a meta-model called \textit{STMeta}, which can consider multiple generalizable spatial and temporal knowledge identified by STAnalytic from literature.

\begin{figure*}[t]
	\centering
	\includegraphics[width=.85\textwidth]{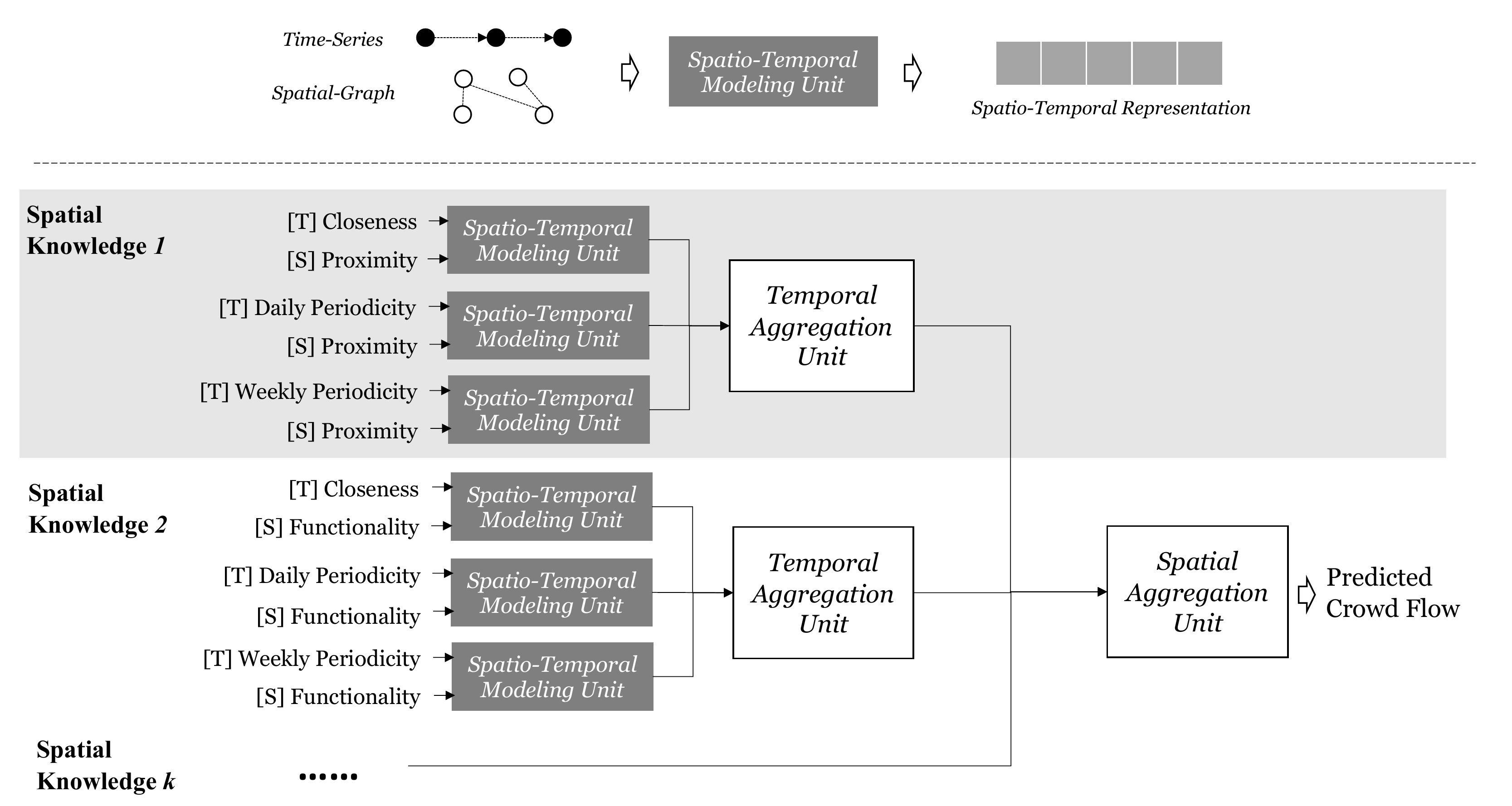}
	\caption{STMeta: a meta-model for spatio-temporal traffic prediction}
	\label{fig:model}
\end{figure*}
\begin{figure}[t]
	\centering
	\includegraphics[width=.9\linewidth]{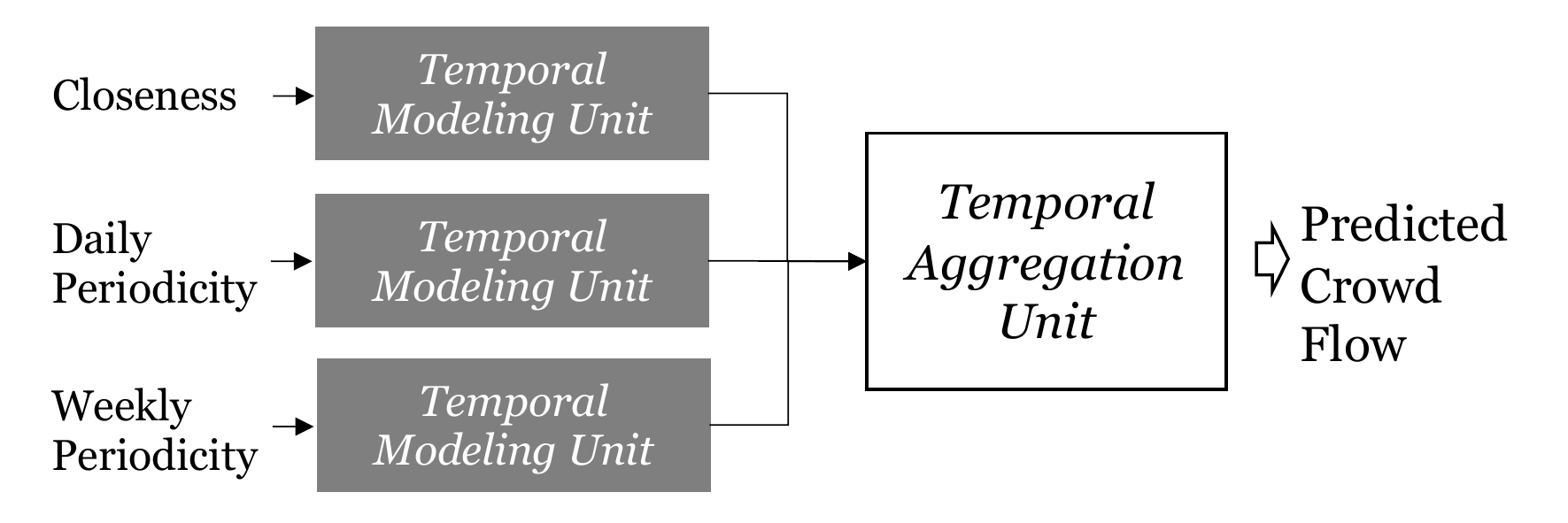}
	\caption{TMeta with temporal factors}
	\label{fig:tmeta}
\end{figure}

\subsection{Design Principles}

Before describing the details of STMeta, we first illustrate the key design consideration of STMeta from two aspects:

(1) \textit{From the modeling perspective, we aim to fully leverage the state-of-the-art deep learning techniques}, which can extract latent features and representations very effectively, especially when we can collect a large amount of historical traffic data with advanced IT infrastructures nowadays.

(2) \textit{From the spatio-temporal knowledge perspective, we attempt to consider the representative and generalizable knowledge captured by STAnalytic}, so that the knowledge that has already been validated from literature can contribute to our model.

%We believe that jointly considering the two aspects are very important for a successful traffic prediction model design. \textit{On one hand}, deep learning techniques can be easily overfit and thus the learned models may be hard to generalize to future prediction; encoding the spatio-temporal knowledge into the deep model can indeed serve as effective domain regularizers to avoid the risk of overfit. \textit{On the  other hand}, most classical methods to encode spatio-temporal knowledge may not be able to find complicated patterns from the big data collected today (e.g., ARIMA can only learn linear patterns); then, leveraging deep learning techniques can alleviate this shortcoming by extracting more complex patterns.

Note that STMeta is a `\textit{meta-model}' of model: many components of STMeta are not restricted to specific types of learning techniques, but can be implemented by alternative techniques that suit the modeling purpose. Next we will describe the details of STMeta. %For example, when we need to model the temporal dynamics of a time series, there are many proper candidate deep learning techniques, such as recurrent neural units (e.g., LSTM or GRU), attention mechanisms, etc. Then, based on the historical traffic data of a specific application, the best technique can be chosen to serve as the final instantiation of our meta model. Next, we describe our model in more details.

\subsection{Meta-Model Details}
\label{sub:STMeta_details}

The overview of STMeta is in Figure~\ref{fig:model}. The key components include the \textit{spatio-temporal modeling}, \textit{temporal aggregation}, and \textit{spatial aggregation} units. We illustrate them as follows.

\subsubsection{Spatio-Temporal Modeling}

The spatio-temporal modeling part of STMeta includes two key steps. First, we decide which input format is suitable for the temporal and/or spatial knowledge. Second, we need some modeling techniques to extract useful patterns from the temporal and/or spatial knowledge inputs.

%\subsubsection{Multi-Series Temporal Knowledge}
To consider the temporal knowledge including closeness, daily periodicity, and weekly periodicity, our model has constructed \textbf{multiple time-series data} regarding different temporal knowledge as the input. Particularly, for \textit{closeness}, the time-series consists of the traffic data in recent time slots; for \textit{daily} (or \textit{weekly}) \textit{periodicity}, the time-series includes the traffic data in the same time slot of last a few days (or the same weekday-time slot in last a few weeks). We use this input format because it is flexible and can be generalized to other cases. For example, if certain traffic data has bi-weekly/monthly/yearly periodicity patterns, it is easy to implement them into STMeta by adding a corresponding time series.

%\subsubsection{Multi-Graph Spatial Knowledge} 
To consider spatial knowledge, we adopt the \textbf{inter-location relationship graph} as the input because of its generality \citep{geng2019spatiotemporal,chai2018bike} (Details in `Inter-Location Relationship' of last section). Then, for specific STTP scenarios, researchers can build suitable inter-location relation graphs, e.g., proximity and functionality, to encode the useful spatial knowledge.

Given the temporal and spatial inputs, the key component of our model is the \textbf{spatio-temporal unit} that takes both \textit{time-series data} (temporal knowledge) and \textit{graph structures} (spatial knowledge)  into account. Recent deep learning research has developed techniques such as \textit{graph convolutional long short-term memory} (GCLSTM) \citep{chai2018bike} and \textit{diffusion convolutional gated recurrent unit} (DCGRU) \citep{li2017diffusion}.

(1) \textit{GCLSTM}: Graph convolution works on a graph $G = (V, A)$ where $V$ is the vertices and $A$ is the adjacency matrix. Let $L=I-D^{-1/2} * A * D^{-1/2}$ be the normalized Laplace matrix, where $D_{ii}=\sum_jA_{i,j}$ is the diagonal degree matrix. The graph convolution can be computed by Chebyshev approximation \citep{defferrard2016convolutional} : $X'=\sum_{k=0}^K{T_k(L)*X*\theta_k}$
where $T_k(L)$ is the $k$-$th$ Chebyshev polynomial and $\theta_k$ is a parameter matrix. To implement GCLSTM, we modify the input $X_t$ and the hidden state $h_{t-1}$ of the LSTM unit to a graph convoluted version: 
%Figure \ref{fig:one} shows the cell structure of GCLSTM. Let $X \in \mathbb{R}^{N,D}$ represent the graph structure data where $N$ is the number of vertices and $D$ is the feature dimension. We set the hidden state of LSTM to $h,c \in \mathbb{R}^{N,M}$, where $M$ is the number of hidden units. The first dimension of $h,c$ is exactly the same as the number of vertices such that GCLSTM can keep cell states for all the stations at the same time. The computing process of GCLSTM is shown as follow:
\begin{align}
	& X_t'=\sum_{k=0}^K{T_k(L)*X_t*\theta_k} \\
	& h_{t-1}'=\sum_{k=0}^K{T_k(L)*h_{t-1}*\theta_k}
%	&f=\sigma(W_f[X_t'; h_{t-1}']+b_f) \\
%	&i=\sigma(W_i[X_t';h_{t-1}']+b_i) \\
%	&o=\sigma(W_o[X_t';h_{t-1}']+b_o) \\
%	&\tilde{c}=tanh(W_c[X'_L;h'_L]+b_c) \\
%	&c_t=f \circ c_{t-1} + i \circ \tilde{c} \\
%	&h_t=o \circ tanh(c_t)
\end{align}
The graph convolution part in GCLSTM can consider the spatial information while the LSTM part can take temporal information.
%where $W_f,W_i,W_o,W_c \in \mathbb{R}^{D+M, M}$, $b_f,b_i,b_o,b_c \in \mathbb{R}^{M}$ are parameter matrices, $[;]$ is the concatenating operation, $\circ$ is the Hadamard product. Suppose the input series is $[X_1,X_2,...X_{t}]$, we then take $h_{t}$ as the final output of GCLSTM.

(2) \textit{DCGRU}: From the design concept, DCGRU is very similar to GCLSTM. That is, DCGRU also re-designs the recurrent neural unit by considering node relations in a graph. The difference is that it uses GRU instead of LSTM, and also replaces the graph convolution operation with the diffusion convolution one. 

%In fact, the diffusion convolution is equivalent to graph convolution under certain conditions \citep{li2017diffusion}, and whether GRU or LSTM performs better often depends on cases. Hence, GCLSTM and DCGRU are alternative implementations of the spatio-temporal unit. %Leveraging which one as the final implementation can depend on real applications.

%\textit{Implementation of Spatio-Temporal Unit}: 
%Besides of these techniques that have been used in literature, we also propose to use a \textit{graph convolutional long-short term memory (GCLSTM)} layer to jointly consider temporal and spatial knowledge in the work, which can be easily adapted to extract spatio-temporal patterns from any spatial location-wise graph with time-series data. Moreover, in previous literature when graph convolution is combined with recurrent network techniques \citep{chai2018bike,geng2019spatiotemporal}, the graph convolution and recurrent operations are often stacked one by one, e.g., first conducting graph convolution on a spatial location-wise graph, and then applying gated recurrent unit (GRU) on a time-series of convoluted graph representations \citep{geng2019spatiotemporal}. That is, while both spatial and temporal factors are considered, in each network layer, only the spatial or temporal factor is considered (Figure XXX).

%In comparison, our GCLSTM layer can take directly the 

\subsubsection{Spatial and Temporal Knowledge Aggregation}

As shown in Figure~\ref{fig:model}, after the spatio-temporal modeling unit, STMeta will first conduct \textit{temporal aggregation} to integrate multi-temporal knowledge, and then \textit{spatial aggregation} to merge multi-spatial knowledge. Similar to the spatio-temporal model unit, there are many candidate deep learning techniques that can be used for temporal and spatial aggregation.

(1) \textit{Graph Attention Layer}. After firstly being introduced by Google \citep{vaswani2017attention}, the attention mechanism has quickly been popular in the deep network structure design. One of its main usages is to aggregate multiple features into an integrated one by learning weights for each feature. %Then, for our learned representations from different temporal or spatial knowledge, we can leverage the attention mechanism to merge them. 
Particularly, we introduce a method of using the \textit{graph attention layer} (GAL) \citep{velivckovic2017graph} to merge multiple features.
The input of GAL is a set of node features $\mathbf h=[h_1,h_2,...,h_N]$, $h_i \in \mathbb{R}^F$, each node being the feature learned from one specific temporal or spatial knowledge (e.g., for three temporal features as closeness, daily and weekly periodicity, $\mathbf h$ has three elements). We first conduct linear transform parametrized by a shared weight matrix $W \in \mathbb{R}^{F,F'}$, and then perform self-attention on each node using a shared attention parameter $a \in \mathbb{R}^{2F'}$:
\begin{equation}
	e_{i,j}= a \cdot [h_i \cdot W; h_j \cdot W]
\end{equation}
where $e_{i,j}$ represents the importance of $h_j$ to $h_i$. To make the attention coefficient comparable, we normalize them using softmax:
\begin{equation}
\alpha_{i,j} = softmax_j(e_{i,j})
\end{equation}
The output of GAL at $h_i$ can then be represented as 
\begin{equation}
h'_i=\sigma(\sum_j{\alpha_{i,j}*(h_j \cdot W)})
\end{equation}
where $\sigma$ is the activation function and we use leaky RELU \citep{velivckovic2017graph}. To make the self-attention process more robust, we add the multi-head mechanism into the model \citep{vaswani2017attention} ($M$ is the number of multi-head):
\begin{equation}
h'_i= \frac{1}{M} \sum_{m=1}^{M}\sigma(\sum_j{\alpha^m_{i,j}*(h_j \cdot W^m)})
\end{equation}
As each node has its own feature after GAL, we then use an average pooling layer to aggregate $h'_i$ into a feature vector:
\begin{equation}
	\bar h=AveragePooling(h'_i)
\end{equation}
where $\bar h$ is the final GAL aggregated feature representation from a set of features $\mathbf h$.

(2) \textit{Concatenation with Dense Layer}. Another widely used technique in deep learning to combine multiple features is concatenation. Then, dense layers can be applied to the concatenated features so as to extract useful representations for the target task:
\begin{equation}
	\hat h = Dense([h_1;h_2;...;h_N])
\end{equation}

For the temporal or spatial aggregation unit in Figure~\ref{fig:model}, either of the above two techniques can be selected.

\subsubsection{Combing Together}

With the spatio-temporal modeling unit and the temporal and spatial aggregation unit, we can then implement concrete STTP approaches following STMeta. Particularly, after spatial aggregation, we can simply stack several dense network layers on the aggregated spatio-temporal representations to predict the future traffic.

\subsubsection{TMeta: Considering Only Temporal Factors}

It is worth noting that, following the design principle of STMeta, we can have a simplified version by considering only temporal factors, called \textit{TMeta} (Figure~\ref{fig:tmeta}). The temporal modeling unit can be implemented by LSTM~\citep{gers1999learning} or GRU~\citep{chung2014empirical}. Later in the benchmark experiments, we will also check how TMeta performs so as to investigate how much improvement can be brought into practice by encoding spatial knowledge.

\subsubsection{Comparison with Existing Methods}

Compared with the existing methods listed in Table~\ref{tab:existing_work}, STMeta is flexible to incorporate a variety of temporal and spatial factors. In particular, ST-MGCN \cite{geng2019spatiotemporal} has also considered multiple temporal and spatial knowledge factors, but its network structure is different from STMeta in temporal modeling. More specifically, ST-MGCN directly combines the weekly, daily, and closeness historical records into \textit{one} time-series sequence as the input. In comparison, the inputs of STMeta include \textit{three} types of time-series data regarding closeness, daily periodicity, and weekly periodicity, respectively; thus, STMeta introduces a temporal aggregation unit to combine the three temporal patterns. We think that the design of STMeta may help to learn different temporal patterns more easily, as each temporal pattern (closeness, daily, and weekly periodicity) has its own time-series sequence for dedicated learning.

\color{black}

%!TEX root = stmeta.tex

\section{Benchmark Experiment}
\label{sec:evaluation}

%In this section, we evaluate STMeta and a number of state-of-the-art STTP methods with ten datasets in five scenarios. %We first introduce the datasets, experiment settings, and then illustrate the results.

\subsection{Benchmark Datasets}

Ten datasets are used in the experiment, covering five scenarios including \textbf{bikesharing demand} (New York City, Washington D.C., and Chicago), \textbf{ridesharing order} (Xi'an and Chengdu)\footnote{We conducted ridesharing experiments on both regular (grid) and irregular (administrative district) region-based prediction tasks.}, \textbf{metro flow} (Shanghai and Chongqing), \textbf{electric vehicle charging station usage} (Beijing), and \textbf{traffic speed} (Los Angeles and Bay Area). 
The length of a time slot has three different settings, 15/30/60 minutes. Our task is to predict the traffic information at the next time slot. \color{blue} Due to the page limitation, dataset details are described in the online appendix. %The city area and the locations for each dataset are shown in Figure~\ref{fig:cities}. The dataset statistics are listed in Table~\ref{tab:datasets}.
\color{black}

\subsection{Experiment Settings}

\subsubsection{Hardware and Training Configurations}

The configurations of the experiments are as follows:

\textbf{Hardware}. Our experiment platform is a computation server with AMD Ryzer 9 3900X CPU (12 cores @ 3.80 GHz), 64 GB RAM, and Nvidia RTX 2080Ti GPU (11GB).

\textbf{Train/Validation/Test Split}. We choose the last 10\% duration in each dataset as test data, the 10\% data before the test for validation. The prediction granularity is set to three settings including 15, 30, and 60 minutes for all datasets. %In the data preprocessing stage, following a common practice \citep{geng2019spatiotemporal}, we remove the stations (or grids) with very small average daily traffic (smaller than 1 in our experiments), since predictions for these low-frequency targets are often not important in real life applications. 

\textbf{Training Stopping Criteria}. 
In training, as different datasets and methods need a varying number of epochs for convergence, we thus leverage \textit{t-test} for the training stopping criteria instead of setting a fixed number of epochs. Particularly, we divide the validation loss of recent epochs (the number of recent epochs is called \textit{early stop patience}) into two halves. For example, if the early stop patience is 100, then the two halves are last 1-50 epochs and last 51-100 epochs. Then, we perform the independent sample t-test on the validation losses of the two halves. When the p-value is smaller than a threshold (we set it to 0.1), the validation losses of the two halves are statistically different, and thus we continue training. The setting of early stop patience is critical, as a small value may stop training too early when the model is still unstable, while a large value may let early stop become useless and lead to overfitting. With trial-and-error, the early stop patience values of the bikesharing, ridesharing, metro, and EV datasets are set to 200, 1000, 400, and 400, respectively, in our experiments.

\subsubsection{Spatio-Temporal Factors in Consideration} %Following the analysis results of the literature in Section~\ref{sub:st_factors}, 
We consider the following spatio-temporal factors.

	\textbf{Temporal}: We consider temporal \textit{closeness}, \textit{daily}, and \textit{weekly periodicity} in our experiments. For each temporal factor in STMeta, we build one time-series data.
	
	\textbf{Spatial}: For all the datasets, we consider the spatial information including proximity and functionality. For \textit{proximity}, we build a graph by linking locations whose distance is smaller than a threshold. For \textit{functionality}, we use the Pearson correlation of the historical traffic data between two locations to indicate their functionality similarity, and link locations with a large correlation. For ridesharing and bikesharing cases, we also build an \textit{interaction} graph by linking the two locations with frequent interactions (many users go from one location to another) \citep{chai2018bike}; for metro, we construct a \textit{same-line} graph by linking the stations in the same line together. The thresholds for building the graphs are shown in Table~\ref{tab:thresholds}. We select these thresholds so that the average number of connections for each node is around 20--30\% of the total number of nodes, which performs well in our experiments.

%\subsubsection{Graph settings}

%Follow the typical setting in GCN\citep{defferrard2016convolutional}, we build undirected and unweighted graphs for each dataset. Nodes with high correlation are linked together. Four kinds of graphs are designed to represent different information, which are distance-graph(DG), correlation-graph(CG), interaction-graph(IG) and line-graph(LG). DG, CG and IG vary in correlation computing metrics and they build links between nodes if their correlation is higher than a certain threshold. Regarding the correlation computing matrics, DG uses geography distance, CG uses pearson correlation and IG uses the number of interactions between tow nodes(e.g. the numbe of bike tracks between two stations). We set different thresholds for these three graphs on different datasets and table \ref{tab:thresholds} shows the detail setting. IG is only built on bike and didi datassets, since the interaction information is not available in the other two datasets. The LG is special designed for the crowd data with certain tracks, e.g. metro data, where the stations on the same route will be linked together. We inplement LG on the metro dataset.

%To get practical values for the thresholds, we performed a parameter search experiment on DiDi data and found that when the average number of connections for each node is around 20\%~30\% of the total number of nodes, the model's performance is better. Based on this discovery, we set different thresholds for these three graphs on different datasets and table \ref{tab:thresholds} shows the detail setting.

\begin{table}[t]
	\caption{Inter-location relationship graph thresholds. For proximity (prox.) graph, it is distance (meter); for functionality (func.) graph, it is Pearson correlation; for interaction (inte.) graph, it is \#records per month.}
	\label{tab:thresholds}

		\begin{center}	
		\resizebox{1\textwidth}{!}{
			\begin{tabular}{lccccc}
				\toprule
           & \textbf{Bikesharing} & \textbf{Ridesharing} & \textbf{Metro} & \textbf{EV} & \textbf{Speed} \\
\midrule
\textbf{Prox.} & 1,000 & 7,500 & 5,000 & 1,000 & 5,500\\
\textbf{Func.}   & 0    & 0.65 & 0.35 & 0.1 & 0.73 (LA)/0.63 (Bay) \\
\textbf{Inte.} & 40  & 30 & --- & --- & --- \\
				\bottomrule
			\end{tabular}}
		\end{center}
	
\end{table}

\subsubsection{STMeta Implementations} 
As shown in Table~\ref{tab:stmeta-version}, we implement three variants of STMeta by changing the techniques used in spatio-temporal modeling and aggregation units (see Section~\ref{sub:STMeta_details}). %, also listed in Table~\ref{tab:stmeta-version}.

Specifically, GCLSTM and DCGRU both contain one layer and 64 hidden units; GAL is set to include two heads and 64 hidden units. After temporal and spatial aggregations, we further stack two dense layers with 64 hidden units to generate the prediction output. We choose the optimization algorithm as ADAM \citep{kingma2014adam}; the learning rate is 1e-5. %To explore different structures and show the flexibility of the model, we implemented three different structure shown in table ~\ref{tab:amulti-version}. The different labels(e.g. T/C) shown in table ~\ref{tab:result-all} represent different components of temporal or graph features, T/CPT means modeling hourly, dayly and weekly feature together, T/C denote only model hourly feature. G/C, G/D, G/DCI represent model correlation graph, distance graph and model three graphes together(distance, crrelation and interaction) respectively.
\color{black} 
We also implement TMeta (only temporal factors) with LSTM~\citep{gers1999learning}.

%The remaining stations (or grids) are listed in each dataset statistic table. 

\subsubsection{Benchmark Approaches} We implement a number of benchmark approaches in the literature. These approaches fall into two types, the first only considers the temporal factors and the second considers both temporal and spatial factors.

Approaches with \textbf{only temporal} factors:

\begin{itemize}
	\item \textbf{HM} (Historical Mean) predicts the traffic according to the mean value of the historical records.  We implement two variants of the HM algorithms. The first considers only the recent time slots (closeness), denoted as \textit{HM (TC)}. \color{black} The second averages recent time slots (closeness), the historical records in the same time of last day (daily periodicity) and last week (weekly periodicity), thus considering multiple temporal factors, denoted as \textit{HM (TM)}.
	
	\item \textbf{ARIMA} \citep{williams2003modeling} is a widely used time series prediction model, which mainly considers temporal closeness.
	
	%\item \emph{HMM} \citep{chen2016predicting}: Hidden markov model is also feasible and effective in traffic flow prediction.
	
	\item \textbf{GBRT} (Gradient Boosted Regression Trees)  \citep{li2015traffic} can be applied to predict the traffic for each location. Our GBRT implementation uses historical traffic data as features, not only from recent time slots, but also from last day and last week. Hence, GBRT considers multiple temporal factors.
	
	\item \textbf{XGBoost} \citep{chen2016xgboost} is similar to GBRT, another widely used tree-based machine learning model.
	
	\item \textbf{LSTM} \cite{ma2015long} neural networks take the recent time-slot traffic data as inputs (closeness) and predict the future.
\end{itemize}

Approaches with \textbf{both temporal and spatial} factors:
\begin{itemize}
	\item \textbf{ST-ResNet} \cite{zhang2017deep} leverages residual convolution networks to consider spatial proximity and also considers temporal closeness, daily and weekly periodicity simultaneously. Note that ST-ResNet can only work for grid-based traffic prediction (ridesharing). 

	\item \textbf{DCRNN} \cite{li2017diffusion}  combines diffusion convolution and recurrent networks for capturing spatio-temporal features. The original DCRNN model only considers the temporal closeness and the spatial proximity graph.

	\item \textbf{STGCN} \cite{yu2018spatio} combines graph convolution and gated convolution units to catch spatial and temporal features, respectively. It considers the temporal closeness and spatial proximity graph.
	
	\item \textbf{GMAN} \cite{zheng2020gman} leverages attention mechanisms to model both temporal closeness and spatial proximity patterns for traffic prediction.
	
	\item \textbf{Graph-WaveNet} \cite{wu2019graph} designs a data-driven graph convolution method for adaptively learning spatial knowledge in addition to proximity. For temporal knowledge, it considers only recent traffic data.

	\color{black}
	
	\item \textbf{ST-MGCN} \cite{geng2019spatiotemporal} captures multiple spatial relations with graph convolutions. It also considers temporal closeness, daily and weekly periodicity. Not like STMeta, ST-MGCN concatenates the inputs regarding different temporal factors into one time-series data.

	\item \textbf{ARGCN-CDW} purely relies on data-driven graph convolution methods for extracting spatial knowledge \cite{bai2020adaptive}. Meanwhile, it considers temporal closeness, daily and weekly periodicity same as ST-MGCN.\footnote{The original ARGCN \cite{bai2020adaptive} considers only temporal closeness, but it is hard to converge in our experiments. By modifying its input as the time-series data combined by closeness, daily and weekly periodicity (same as ST-MGCN), the performance of ARGCN is much improved. We denote this model as ARGCN-CDW.}
	\color{black}

\end{itemize}

\begin{table}[t]	
	\caption{STMeta and TMeta implementations. (ST-Unit: Spatio-Temporal Unit; TA-Unit: Temporal Aggregation Unit; SA-Unit: Spatial Aggregation Unit)}
	\footnotesize
	\label{tab:stmeta-version}
	\vspace{-1em}
		\begin{center}
			\resizebox{.8\textwidth}{!}{
			\begin{tabular}{lccc}
				\toprule
				 & \textit{ST Unit} & \textit{TA Unit} & \textit{SA Unit}\\
				\midrule
				\textbf{STMeta-GCL-GAL} & GCLSTM & GAL     & GAL \\
				\textbf{STMeta-GCL-CON} & GCLSTM & Concatenation & GAL \\
				\textbf{STMeta-DCG-GAL} & DCGRU  & GAL     & GAL \\
				\midrule
				\textbf{TMeta-LSTM-GAL} & LSTM & GAL & --- \\
				%\textbf{TMeta-GRU-GAL} & GRU & GAL & - \\
				\bottomrule
			\end{tabular}}
		\end{center}
\end{table}

\begin{table*}[t]
	\caption{60-minute prediction error. The best two results are highlighted in bold, and the top one result is marked with `*'. (TC: Temporal Closeness; TM: Multi-Temporal Factors; SP: Spatial Proximity; SM: Multi-Spatial Factors; SD: Data-driven Spatial Knowledge Extraction)}
	\label{tab:result-all-60}
	\begin{center}
		
		\resizebox{\textwidth}{!}{
			\begin{tabular}{lcccccccccccccc}
				\toprule
				& \multicolumn{3}{c}{\textbf{Bikesharing}} & \multicolumn{4}{c}{\textbf{Ridesharing}} & \multicolumn{2}{c}{\textbf{Metro}} & \textbf{EV} & \multicolumn{2}{c}{\textbf{Speed}} & \multicolumn{2}{c}{\textbf{Overall}} \\
				\cmidrule(lr){2-4} \cmidrule(lr){5-8} \cmidrule(lr){9-10} \cmidrule(lr){11-11} \cmidrule(lr){12-13} \cmidrule(lr){14-15}
				& \textit{NYC} & \textit{CHI} & \textit{DC} & \textit{XA-gr.}  & \textit{CD-gr.} & \textit{XA-di.} & \textit{CD-di.} & \textit{SH} & \textit{CQ} & \textit{BJ} & \textit{LA} & \textit{Bay} & \textit{AvgNRMSE} &\textit{WstNRMSE}\\
				\midrule
				\textbf{Temporal} &&&&&&&&&&&&&&\\
				\emph{HM (TC)}           &5.814 &	4.143 &	3.485 &	10.136 &	14.145  &52.610 &	74.212  & 824.94 &	673.55 &	1.178 &	12.303& 	5.779  & 2.597 &	7.106 
				\\
				\emph{ARIMA (TC)}           & 5.289 & 3.744 & 3.183 & 9.475 & 13.259 &47.794 &	65.725 & 676.79 & 578.19 & 0.982 & 11.739 &5.670 & 2.297 &6.100\\
				\emph{LSTM  (TC) }     		& 5.167 & 3.721 & 3.234 & 9.830 & 13.483 &43.962 &	51.355 & 506.07 & 322.81 & 0.999 & 10.083 &4.777 & 1.882 &	3.535 \\
				\emph{HM (TM)}       		& 3.992 & 3.104 & 2.632 & 6.186 & 7.512  & 27.821 &	30.917 & 172.55 & 119.86 & 1.016 & 10.727 &4.018 & 1.180 &	1.265 \\
				\emph{XGBoost (TM)}  		& 4.102 & 3.003 & 2.643 & 6.733 & 7.592  & 27.745 &	30.347 & 160.38 & 117.05 & 0.834 & 10.299 &3.703& 1.146 &	1.235 \\
				\emph{GBRT (TM)}	   		& 4.039 & 2.984 & 2.611 & 6.446 & 7.511  & 25.654 &	27.591 & 154.29 & 113.92 & 0.828 & 10.013 &3.704& 1.111 & 1.202\\
				\emph{TMeta-LSTM-GAL (TM)}     		& 3.739 & 2.840 & 2.557 & \textbf{5.843} & 6.949 &\textbf{23.024}  &	\textbf{25.264*} & 163.31 & 102.86 & 0.840 & \textbf{8.670*} &3.616& 1.047 &1.141\\
				\midrule
				\textbf{Temporal \& Spatial} &&&&&&&&&&&& \\
				\emph{DCRNN (TC+SP)} 		& 4.187 & 3.081 & 3.016 & 8.203 & 11.444 & 39.028 &	48.393 & 340.25 & 122.31 & 0.989 & 11.121 &6.920& 1.544  &2.376\\
				\emph{STGCN (TC+SP)} 		& 3.895 &	2.989 &	2.597 &	6.150  & 7.710  &	23.916 	& 28.940   & 187.98 &	106.16 &	0.859 &	10.688 &	3.472 & 1.121 &	1.313 \\
				\emph{GMAN (TC+SP)} 		& 4.251 &	2.875 &	2.530 &	7.099 &	13.351   &	28.086 &	30.741   & 193.39  &	117.52 &	0.949 &	10.012 &	3.846   & 1.251 	& 1.947 \\
				\emph{Graph-WaveNet (TC+SP+SD)} 		& 3.863 &	2.812 &	\textbf{2.403*} &	6.541 &	8.162   &	24.101 &	29.406   & 186.82 &	102.75  &	0.930 &	9.463 &	4.135  & 1.129 &	1.305 
				\\
				\emph{ST-ResNet (TM+SP)}   	&  ---  &  ---  &  ---  & 6.075 & 7.155  &  ---   &  --- & --- & ---   &  ---  & --- &--- & --- & ---\\
				%\emph{ASTGCN (TM+SP)} \\
				\emph{ST-MGCN (TM+SM)}& 3.723 & 2.904 & 2.518 & 5.878 & 7.067  & 30.413& 	26.014 & 159.52 & 104.87 & 0.827 & 10.798 &\textbf{3.486}& 1.094&1.324\\
				\emph{AGCRN-CDW (TM+SD)} 		& 3.795 &	2.935 &	2.580 &	8.835 &	10.275    &	35.801 	& 38.093   & 658.12 &	287.41  &	0.844 &	10.728 &	\textbf{3.381*}   & 1.688 	& 4.596 \\
				\emph{STMeta-GCL-GAL (TM+SM)}		& \textbf{3.518} & \textbf{2.695} & 2.405 & 5.871 & \textbf{6.858*}  & 24.127 &	25.669 & 153.17 & \textbf{97.87}  & 0.831 & \textbf{8.834} &3.514& \textbf{1.024} &\textbf{1.070*}\\
				\emph{STMeta-GCL-CON (TM+SM)} 	& \textbf{3.507*} & 2.739 & \textbf{2.404} & \textbf{5.829*} & \textbf{6.873}  & 23.244 &	\textbf{25.296} & \textbf{149.05}  & 106.41  & \textbf{0.807} & 9.147 &3.552 & 1.027 & 1.123 \\
				\emph{STMeta-DCG-GAL (TM+SM)}  & 3.521 & \textbf{2.652*} & 2.423 & 5.908 & 6.904  & \textbf{22.979*} &	27.217  &\textbf{143.18*}  & \textbf{94.78*} & \textbf{0.803*}& 8.993 &3.500 & \textbf{1.015*} & \textbf{1.077} \\
				\bottomrule
		\end{tabular}}
	\end{center}
\end{table*}

 The hyperparameters of the above benchmark approaches follow the default settings as their original papers.\footnote{ We have also tested some other approaches in Table~\ref{tab:existing_work}, such as ASTGCN \cite{guo2019attention} and STSGCN \cite{song2020spatial}; however, they are hard to converge under certain experiment scenarios.} \color{black}

\subsubsection{Metric}
We use RMSE (root mean square error) to report the prediction error, as this is the most widely used metric in almost all the STTP research \cite{zhang2017deep,liang2018geoman,geng2019spatiotemporal,chai2018bike,Zhang0QLYL18}.\footnote{We have also tested methods in MAE (Mean Absolute Error), and the results are consistent with RMSE. For clarity, we just report RMSE.} 
We also compute two aggregation scores to indicate the overall performance (i.e., generalizability) of an approach $x \in \mathcal X$ ($\mathcal X$ is the set of all approaches) among all the datasets $\mathcal D$:
\begin{equation}
	\textit{AvgNRMSE}_x  = avg_{d\in \mathcal D}(\frac{RMSE_{x, d}}{min_{x'\in \mathcal X}(RMSE_{x',d})})
\end{equation}
where $RMSE_{x,d}$ is the RMSE of approach $x$ in dataset $d$. The meaning of the score is the average of the normalized RMSE (normalized by the smallest RMSE for every dataset $d$). %Then, the smaller the score is, the better the approach performs. 
Ideally, if an approach can perform the best among all datasets, then \textit{AvgNRMSE} is $1$. 
\begin{equation}
	\textit{WstNRMSE}_x  = max_{d\in \mathcal D}(\frac{RMSE_{x, d}}{min_{x'\in \mathcal X}(RMSE_{x',d})})
\end{equation}
If a method is generalizable, its \textit{WstNRMSE} should also be close to $1$, i.e., in the worst dataset, its performance is still near the best one.

\subsection{Experiment Questions}

We aim to answer the following questions by experiments.

\textbf{Q1. Model Generalizability}. \textit{Can STMeta demonstrate better generalizability than state-of-the-art models?} We deem that a model is more generalizable, if it can achieving lower AvgNRMSE and WstNRMSE.

\textbf{Q2. Knowledge Generalizability}. \textit{Which temporal or spatial knowledge factors are more generalizable across different datasets?} A generalizable factor means that a model considering the factor would exhibit significantly lower AvgNRMSE and WstNRMSE than a model not considering the factor.

\subsection{Model Generalizability Results}

Table~\ref{tab:result-all-60}, \ref{tab:result-all-30} and \ref{tab:result-all-15} (Appendix) show our evaluation results for the 60, 30, and 15-minute time slots. For each approach, we mark in the bracket which temporal and/or spatial factors are considered following STAnalytic.  The best two results for each dataset are highlighted in bold. 

First, we highlight the generalization effectiveness of our STMeta variants over different datasets. From the overall results in Table~\ref{tab:result-all-60}, \ref{tab:result-all-30} and \ref{tab:result-all-15}, we can see that the three different variants of STMeta are usually ranked among the top methods. Particularly, the top method of the three experiments with different time slot lengths is always \textit{STMeta} variants regarding both \textit{AvgNRMSE} and \textit{WstNRMSE}, which verifies the better generalizability of our proposed meta-model STMeta compared to other benchmark approaches. %For example, \textit{STMeta-GCL-GAL} outperforms the best benchmark approach \textit{ST-MGCN} by reducing around 5\% of prediction error on average. 
This indicates that our STMeta has the potential to serve as a generalizable STTP meta-model to guide future approach design for a variety of scenarios. Regarding different variants, which one is better depends on the application scenario. This also matches our expectation as we believe that the concrete component implementations in STMeta can have alternative choices (e.g., GCLSTM or DCGRU for the spatio-temporal modeling unit) depending on the scenarios. In the future, if new modeling techniques are proposed, it is possible to implement new STMeta variants. 

%It is also worth noting that, the simplified version of STMeta with only temporal factors, i.e., \textit{TMeta}, performs surprisingly well, ranked fourth in 60/30-minute experiments regarding \textit{AvgNRMSE} (only behind three STMeta variants), even better than some approaches considering both temporal and spatial factors. This is perhaps because if there exists a long duration of data, it is possible to learn the traffic dynamic pattern only from a location's own historical data. This inspires: (i) \textbf{Spatial knowledge needs more careful design to take effect}; (ii) \textbf{When designing an STTP model for a new scenario, one can first try TMeta }as it is easy to implement and consistently performs well in various scenarios.

%\subsubsection{Modeling Technique Impact on Prediction}
%\label{sub:exp_modeling}
We also investigate the generalizability of benchmark modeling techniques. %, i.e., studying how different modeling techniques impact the prediction accuracy across different datasets. 
To reduce the impact of knowledge variation on our analysis, we compare different approaches with the same spatial and temporal knowledge factors. For `TM' approaches, we find that their overall performance \textit{AvgNRMSE} ranges from 1.047/1.047/1.086 (TMeta-LSTM-GAL (TM)) to 1.180/1.235/1.246 (HM (TM)) for 60/30/15-minute time slot. On one hand, this validates that advanced modeling techniques can actually improve the model generalizability toward diverse datasets (e.g., state-of-the-art deep learning techniques of TMeta compared to the naive historical mean of HM). On the other hand, it reveals that the improvement brought by modeling technique is around 10-20\%, much smaller than the improvement of the extra temporal knowledge (e.g., HM (TC) is worse than HM (TM) by more than 100\% in the 60-min experiment). %We also analyze the `TM+SM' approaches, the conclusion is similar: advanced deep modeling techniques that we have included in STMeta can outperform the benchmark approaches such as ST-MGCN, which shows better generalizability; however, the improvement is not very large, limited to be around 5\%.

\begin{table*}[t]
	\footnotesize
	\caption{Temporal knowledge evaluation (60-min). The model is STMeta-DCGRU-GAL. The best result is in bold.  The training/inference time is shown in the brackets. \color{black} (C: Closeness, D: Daily Periodicity, W: Weekly Periodicity)} 
	\label{tab:result-temporal}
	\begin{center}
		
		%\resizebox{.9\textwidth}{!}{
			\begin{tabular}{lccccc}
				\toprule
				& \multicolumn{1}{c}{\textbf{Bikesharing}} & \multicolumn{1}{c}{\textbf{Ridesharing}} & \multicolumn{1}{c}{\textbf{Metro}} & \textbf{EV} & \textbf{Speed}\\
				& \textit{NYC}  & \textit{CD-grid} & \textit{SH} & \textit{BJ} & \textit{LA}\\
				\midrule
				\emph{C}&4.205 (14.22h/10.53s)&8.380 (6.72h/0.11s)&328.81 (7.12h/0.12s)&3.423 (5.09h/0.20s)&10.672 (2.16h/0.19s)\\
				\emph{CD}&3.631 (28.37h/21.54s)&7.517 (9.23h/0.22s)&\textbf{140.67} (10.47h/0.24s) & 0.809 (7.22h/0.40s) & 10.117 (3.35h/0.35s)\\
				\emph{CDW}&\textbf{3.521} (34.46h/27.74s)&\textbf{6.920} (10.03h/0.28s)&143.18 (12.46h/0.31s)&\textbf{0.803} (7.93h/0.52s) & \textbf{8.993} (3.72h/0.45s)\\
				\bottomrule
			\end{tabular}
			%}
	\end{center}
\end{table*}

Another interesting observation is the traditional machine learning model GBRT performs competitively to the state-of-the-art deep learning models in the 15-minute experiment (Table~\ref{tab:result-all-15}). In particular, \textit{AvgNRMSE} of GBRT in 15-minute prediction is 1.097 (i.e., GBRT is only worse by 9.7\% compared to the best model on average). \color{blue} More specifically, in the ridesharing-CD (grid) experiment, GBRT is worse than the best model only by 3.8\%, and outperforms many deep models such as Graph-WaveNet and STGCN. \color{black} %As our performance comparison runs on \color{blue} various benchmark scenarios, \color{black} we do not believe that it happens only because of randomness. 
Considering the computation resources (computation time), we deem that for STTP with short-length time slots, classical machine learning models may be computation-efficient, i.e., getting relatively good accuracy with significantly lower computation resources, than deep learning methods. This is perhaps because the traffic data relations between continuous time slots become more obvious with the decrease of time slot duration, and classical methods can already capture the relations well.

In addition, we find that the data-driven spatial knowledge modeling (the methods with `SD') performs not stable across different datasets. For instance, AGCRN-CDW (TM+SD) has achieved the smallest prediction error in Speed-Bay (60-minute); however, it performs very poorly in the Metro scenarios (60-minute). Occurring such a large difference is perhaps because extracting spatial knowledge purely from data is of distinct difficulties between datasets. While we believe that data-driven spatial pattern extraction is promising, adopting it in practice still needs caution, as the performance is highly application-dependent.

\color{black}

In summary, by comparing STMeta to various benchmark approaches, we validate its superiority on the generalizable effectiveness over different datasets. Meanwhile, for STTP with short-length time slots, traditional machine learning models, such as GBRT, can also be a good selection thanks to its high computation efficiency and relatively good accuracy. %In addition, as STMeta can be instantiated to incorporate different types of knowledge factors, it can also serve as an appropriate base model to evaluate the knowledge factors' generalizability, which we discuss in the next section.

\subsection{Knowledge Generalizability Results}

Here, we analyze the results to see how various knowledge factors impact the prediction. Overall, we observe that the best series of approaches regarding knowledge are the ones that consider the largest number of knowledge factors, i.e., the approaches marked as `\textit{TM+SM}' (multi-temporal and multi-spatial factors). This in fact supports the hypothesis that we have made when we analyze the existing research approaches with STAnalytic (Section~\ref{sub:analyze_existing}), i.e., \textit{the more knowledge, the better prediction}.
%\footnote{While worse than STMeta variants, ST-MGCN~\citep{geng2019spatiotemporal} is the best among existing benchmark approaches.} 
Next, we present a detailed analysis regarding various temporal and spatial factors.

\subsubsection{Temporal Knowledge}

Regarding the temporal knowledge factors, we first investigate the prediction performance of the `temporal' approaches in Table~\ref{tab:result-all-60}/\ref{tab:result-all-30}/\ref{tab:result-all-15}. According to the results, considering more temporal knowledge would improve prediction significantly.  For example, in 60-minute prediction (Table~\ref{tab:result-all-60}), \textit{HM~(TM)}'s \textit{AvgNRMSE} = 1.190; in comparison, \textit{HM~(TC)}'s \textit{AvgNRMSE} = 2.597, which is much worse than \textit{HM (TM)}. \color{black} This reveals involving multiple temporal factors indeed enhances the prediction accuracy.

To further elaborate on the impacts of different temporal factors, we conduct an experiment with STMeta by removing certain temporal factors. The results are shown in Table~\ref{tab:result-temporal}. We observe that adding periodicity can always enhance the prediction accuracy as human mobility has intrinsic regularity \cite{song2010limits};  meanwhile, the training and inference time also increases a bit. \color{black} Generally, adding both daily and weekly periodicity can bring much improvement, while adding only daily periodicity might be the best (metro, Shanghai). Hence, the temporal periodicity knowledge is often generalizable, while which periodicity performs the best depends on applications.

Besides, with the decrease of the time slot length, the performance gap between `TC' and `TM' methods becomes smaller, indicating that the temporal closeness plays a more important role. For example, in the 60-min experiment, \textit{AvgNRMSE} of LSTM (TC) and TMeta-LSTM-GAL (TM) are 1.882 and 1.047, respectively, leading to a gap of 0.835; in the 15-min experiment, \textit{AvgNRMSE} of LSTM (TC) and TMeta-LSTM-GAL (TM) are 1.332 and 1.086, respectively, indicating a gap of only 0.246.
\color{black}
This observation signifies that the \textit{temporal closeness} factor becomes more important when the time slot length is reduced. This also fits our expectation --- the correlation between the next 15-minute traffic data and the recent 15-minute should be higher than the correlation between the next 60-minute traffic data and the recent 60-minute, as the time difference of the two continuous slots in the former case is much smaller than the latter case.

\subsubsection{Spatial Knowledge}
\label{exp:spatial_knowledge}

\begin{table*}[t]
	
	\footnotesize
	\caption{Spatial knowledge evaluation (60-min). The model is STMeta-DCGRU-GAL. The best result is in bold.  The training/inference time is shown in the brackets. \color{black} (F:Function, P:Proximity, I: Interaction, S: Same-Line)}
	\label{tab:result-spatial}
	\begin{center}
		%\resizebox{.9\textwidth}{!}{
			\begin{tabular}{lccccc}
				\toprule
				& \multicolumn{1}{c}{\textbf{Bikesharing}} & \multicolumn{1}{c}{\textbf{Ridesharing}} & \multicolumn{1}{c}{\textbf{Metro}} & \textbf{EV} & \textbf{Speed}\\
				%\cmidrule(lr){2-2} \cmidrule(lr){3-3} \cmidrule(lr){4-4} \cmidrule(lr){5-5} \cmidrule(lr){6-6} 
				& \textit{NYC}  & \textit{CD-grid} & \textit{SH} & \textit{BJ} & \textit{LA}\\
				\midrule
				\emph{F}&\textbf{3.477} (6.40h/5.03s)&\textbf{6.878} (2.62h/0.06s)&148.25 (2.97h/0.06s)&0.813 (2.93h/0.18s)&10.149 (1.64h/0.19s)\\
				\emph{P}&3.509 (6.48h/5.02s)&6.914 (2.60h/0.07s)&153.98 (2.96h/0.06s) & 0.835 (2.93h/0.18s)&10.168 (1.65h/0.19s)\\
				\emph{I/S}&3.543 (6.40h/5.01s)&6.879 (2.60h/0.06s)&151.38 (2.98h/0.06s)&-&-\\
				\midrule
				\emph{Full}&3.521 (34.46h/27.74s)&6.904 (10.03h/0.28s)&\textbf{143.18} (12.46h/0.31s)&\textbf{0.803} (7.93h/0.52s)&\textbf{8.993} (3.72h/0.45s)\\
				\bottomrule
		\end{tabular}
	%}
	\end{center}
\end{table*}

Similarly, to analyze how spatial knowledge impacts prediction, we remove some spatial factors from \textit{STMeta-DCG-GAL} and then compare it to the original \textit{STMeta-DCG-GAL} (with full spatial knowledge). In particular, for the 60-minute prediction, we implement some variants of \textit{STMeta-DCG-GAL} with partial spatial information: one of the proximity, functionality, and interaction/same-line graphs. Due to the page limitation, we only show one dataset for each scenario.

%Since our \textit{STMeta} implementations consider `TM+SM', so we want to find some approaches considering `TM' and partial spatial information.\footnote{Among existing benchmark approaches, only ST-ResNet considers `TM' and partial spatial information (`SP'); however, it only works for grid-based STTP.} To this end, for the 60-minute prediction, we implement some variants of the best method, \textit{STMeta-GCLSTM-GAL}, with only partial spatial information, i.e., one of the proximity, functionality, and interaction/same-line graphs. 

Results are shown in Table~\ref{tab:result-spatial}. First, among different spatial graphs, the functionality graph performs always the best. In this regard, we deem that it is more generalizable across different datasets. Besides, the metro, EV, and speed scenarios benefit from the integration of multiple spatial knowledge more significantly than the bikesharing and ridesharing scenarios. For example, in ridesharing, STMeta with full spatial knowledge perform worse than the approach with only one type of spatial knowledge in Chengdu;  moreover, training STMeta with full spatial knowledge (i.e., three graphs) costs more than 5 times of the computation hours compared with STMeta with only one spatial graph. Hence, if we blindly add a variety of spatial knowledge graphs into a model, it may degrade both the model effectiveness and efficiency. \color{black} \textbf{This inspires that, while generally considering spatial knowledge can improve prediction, selecting and integrating which specific spatial knowledge should be carefully determined for a specific application}. Otherwise, it is possible that the overall prediction performance is degraded by adding more spatial knowledge.

\subsection{Benchmark Implications}

%Based on the previous analysis and benchmark results, we discuss the implications.

%\subsection{Insights and Implications}

\subsubsection{Knowledge vs. Modeling}

With the popularity of machine learning, especially deep learning, more and more studies focus on applying novel modeling techniques to the STTP model design, while which temporal or spatial knowledge is considered often lacks detailed analysis. However, our benchmark results indicate that, \textbf{for STTP model design, it is probable that which part of knowledge in consideration is more important than which modeling technique in use}. For example, when we consider complete temporal closeness, daily and weekly periodicity together in the naive historical mean method, \textit{HM (TM)} can significantly outperform the advanced deep modeling technique with inadequate temporal knowledge, \textit{LSTM (TC)} that considers only the temporal closeness in all the benchmark scenarios. In a word, regarding the \textit{knowledge} and \textit{modeling techniques}, we believe that researchers and practitioners should make more efforts on the knowledge part, i.e., considering more carefully about the temporal and spatial knowledge that may be suitable for the target STTP scenario.

\subsubsection{Temporal Knowledge vs. Spatial Knowledge}

With the benchmark experiments, we can also compare the relative importance between temporal and spatial knowledge in their generalizability. Overall, we find that purely using temporal knowledge can already achieve competitive generalizability performance across different datasets, compared to the approaches with both temporal and spatial knowledge. For example, TMeta-LSTM-GAL's \textit{AvgNRMSE} ranges from 1.047 to 1.086 for 15/30/60-minute prediction experiments. This indicates that the best model with extra spatial knowledge may help improve only $\sim$5\% beyond temporal knowledge. Specifically, for Bikesharing-CHI (30-minute), TMeta-LSTM-GAL performs the best (tie with STMeta-GCL-CON). This reveals that temporal knowledge can dominate the spatio-temporal factors in some cases. In comparison, as shown in Sec.~\ref{exp:spatial_knowledge}, adding spatial knowledge does not always improve the prediction accuracy. For different datasets, which spatial knowledge contributes more varies significantly. Hence, our benchmark results reveal that \textbf{temporal knowledge is more generalizable than spatial knowledge across different STTP scenarios}.

\color{black}

\subsubsection{Design Guidelines} For the researchers needing to design STTP approaches for their applications, we summarize the following guidelines.

	\textbf{Guideline 0 (most important)}: Always think deeply about which knowledge should be encoded in the model design before focusing on sophisticated modeling tricks. For a targeted STTP scenario, finding a good set of knowledge suitable for the scenario is a prerequisite and fundamental step before designing model details.

	\textbf{Guideline 1 (Temporal Knowledge)}: For temporal knowledge, consider temporal closeness, daily and weekly periodicity together. They are perhaps generalized and robust knowledge for various STTP scenarios, as these are fundamental human activity patterns.\footnote{Other scenario-specific temporal knowledge may also be carefully designed. For example, if we predict traffic in special seasons such as the \textit{Spring Festival Travel Season} in China, yearly periodicity is critical.}

 \textbf{Guideline 2 (Spatial Knowledge)}: For spatial knowledge, it needs more careful consideration. Overall, spatial knowledge is not as generalizable as temporal knowledge, whether it is pre-specified or purely learned from data. \color{black} Hence, instead of simply aggregating all the varieties of spatial knowledge, conduct trial-and-error tests to select the best spatial knowledge combination.

 \textbf{Guideline 3 (STMeta)}: Our STMeta can be used as a meta-model to integrate multiple temporal and spatial knowledge. With STMeta, researchers only need to consider how to find effective spatial knowledge and encode the knowledge into an inter-location relationship graph.

 \textbf{Guideline 4 (TMeta)}: To deal with a new STTP problem, we recommend firstly building a TMeta model with only temporal knowledge as an easy-to-implement baseline. This is because temporal knowledge is more generalized than spatial knowledge, and the computation efficiency of TMeta is better than STMeta.
	
\textbf{Guideline 5 (Traditional Models)}:  For STTP with short-length time slot settings, the traditional machine learning models with only temporal knowledge, such as GBRT (TM), can be a nice option to trade off the prediction accuracy and computation efficiency.
	\color{black}

%!TEX root = stmeta.tex
\section{Related Work}

The STTP problem is useful for many urban computing scenarios including ridesharing demand, public transportation flow, EV charging station usage \citep{geng2019spatiotemporal,chai2018bike,zhang2017deep}. Our paper has investigated some representative studies while there are still many others not mentioned in detail \cite{yao2018deep,ke2017short,Pan2019UrbanTP,Lv2015TrafficFP,Zhang2019FlowPI,yao2019revisiting}. Due to the page and time limitation, we have not yet analyzed these approaches and re-implemented their methods for evaluation. We will extend our analysis and experimental comparison by selecting more studies and add their model implementations into our code repository.%\footnote{We plan to frequently update a version of this work (e.g., on ArXiv), by continuously adding the analysis results of new STTP studies published in the future.} 

There are also some research topics related to STTP:

%\textit{Traffic speed prediction} aims to predict vehicles' speed in road segments with sensing data from road loop sensors or vehicle GPS devices \citep{yu2018spatio,Zheng2020GMANAG,Wu2019GraphWF}. This task is very close to crowd flow prediction as spatial and temporal patterns widely exist. Hence, a natural extension of our meta-modeling and analytic framework is generalization to traffic speed prediction cases. We will consider it in the future work. 

\textit{Sensory time-series processing} focuses on mining time-series data generated by pervasive sensors, e.g., accelerometers, gyroscopes, and
magnetometers \citep{yao2018deep} to facilitate applications such as mobile sensing \citep{Lane2010ASO,zhou2015smiler} and activity recognition \citep{Wang2018DeepLF}. Compared to traffic data, usually the time slot length of the sensory time-series data can be much shorter (e.g., in seconds), and thus the temporal knowledge in consideration can be different. However, we still think that STAnalytic may inspire a similar analysis framework for sensory time-series data. 

\textit{Individual mobility prediction} targets predicting an individual person's future visiting locations \citep{Feng2018DeepMovePH,Gao2019PredictingHM}. While this is individual level and STTP involves mostly aggregation level information, there are many commonalities in the leveraged knowledge for building the prediction model. For example, periodicity is recently added by DeepMove \citep{Feng2018DeepMovePH} into individual mobility prediction and offers a significant improvement. Hence, analyzing the literature on individual mobility prediction following a framework similar to STAnalytic can be interesting and valuable. It is also possible that traffic prediction and individual mobility prediction can learn from each other from such analysis.

\section{Conclusion}

In this paper, we propose an analytic framework, called \textit{STAnalytic}, for investigating and comparing existing models for spatio-temporal traffic prediction (STTP). %Existing research efforts on STTP do not take into account whether a model designed for one specific scenario can be generalized to other scenarios. 
%STAnalytic can help researchers and practitioners to investigate different application-specific STTP models in a unified way, thus making them comparable.
Particularly, with STAnalytic, researchers can investigate and compare what high-level spatial or temporal knowledge is exploited in STTP models, and further justify which model may be generalized to other STTP scenarios. Additionally, we propose a ``model of model'', i.e., the meta-model, called \textit{STMeta}, to flexibly integrate multiple spatio-temporal  knowledge identified by STAnalytic from literature. With ten real-life datasets, we have demonstrated that STMeta can generally work well on various scenarios. 

Lastly, we re-highlight the novel ideas in this research, which may advance the STTP research promisingly:

(1) \textit{Re-framing the fundamental thinking paradigm for STTP research}. From the pioneering deep learning-based STTP research in 2017 \cite{zhang2017deep}, the STTP area becomes enthusiastic about adopting novel machine and deep learning techniques. %While this significantly improves the STTP area up to date, we may slow down and re-think the question, `\textit{Is continuously applying the new learning techniques into STTP still enough to advance the area in the future?}' % Specifically, 
Our research reveals that only putting efforts into techniques may not be enough, and we should put more focus on \textit{which knowledge factors are considered and whether these factors are generalizable}. %This is a fundamental change for the thinking paradigm of STTP research.

(2) \textit{Proposing the first (partially) analytical methodology to compare STTP approaches}. Generally, no one machine learning technique can win all the time, i.e., the machine learning `No free lunch' theorem.\footnote{http://www.no-free-lunch.org/} Most prior STTP papers only rely on experiments over certain datasets to verify performance `\textit{empirically}'. However, whether these approaches can work well for un-tested datasets is still questionable to a certain extent. In comparison, with \textit{STAnalytic}, our research proposes perhaps the first methodology to compare STTP approaches (partial) `\textit{analytically}' by considering their considered knowledge factors. %That is, we can compare approaches with their considered knowledge factors and then justify their performance from the knowledge aspect. %(Sec~\ref{sub:analyze_existing}). %For example, if Approach \textit{A} neglects certain significant spatio-temporal factors compared to Approach \textit{B}, we can confidently deduce that \textit{A} would perform worse without empirical experiments. This is a conceptually new methodology to compare STTP approaches, which we believe can be a useful complement to the empirical experiments for STTP research.

\color{blue}
Our future work would include: (1) We will continue updating our project website (\url{https://github.com/uctb/UCTB}) to cover new STTP models and datasets. (2) We will extend experimental STTP tasks to multi-step prediction. Given an STTP model, various techniques could be applied to achieve multi-step prediction \cite{taieb2015bias}. It would be interesting to investigate whether (i) there is one single technique that dominantly performs the best, or (ii) different STTP models should cooperate with diverse techniques to achieve the best multi-step prediction performance. %(3) We will further quantitatively study how external factors would impact STTP models under various scenarios. 

\color{black}

\section*{Acknowledgment}

This work is supported by NSFC Grant no. 61972008, PKU-Baidu Fund Project no. 2019BD005, and Hong Kong RGC Theme-based Research Scheme no. T41-603/20-R. 

%Beyond the qualitative way to investigate various STTP models, we have constructed a set of large-scale STTP testbed datasets for quantitatively validating the generalizability of STTP models on different scenarios. The dataset covers four scenarios including ridesharing, bikesharing, metro, and electrical vehicle charging in eight cities in China. The experiment on this dataset has verified the general effectiveness of STMeta over the state-of-the-art STTP models by reducing up to 8.2\% prediction error. 

\color{black}
\bibliographystyle{IEEEtran}
\bibliography{crowd_flow}

\newpage

\newcommand{\beginsupplement}{%
	\setcounter{table}{0}
	\renewcommand{\thetable}{S\arabic{table}}%
	\setcounter{figure}{0}
	\renewcommand{\thefigure}{S\arabic{figure}}%
}

\beginsupplement

\onecolumn
\subsection*{Supplementary Tables on Experimental Results}

\begin{table*}[h!]
	\caption{30-minute prediction error. The best two results are highlighted in bold, and the top one result is marked with `*'. (TC: Temporal Closeness; TM: Multi-Temporal Factors; SP: Spatial Proximity; SM: Multi-Spatial Factors; SD: Data-driven Spatial Knowledge Extraction)}
	\label{tab:result-all-30}
	\begin{center}
		
		\resizebox{\textwidth}{!}{
			\begin{tabular}{lcccccccccccccc}
				\toprule
				& \multicolumn{3}{c}{\textbf{Bikesharing}} & \multicolumn{4}{c}{\textbf{Ridesharing}} & \multicolumn{2}{c}{\textbf{Metro}} & \textbf{EV} & \multicolumn{2}{c}{\textbf{Speed}}& \multicolumn{2}{c}{\textbf{Overall}} \\
				\cmidrule(lr){2-4} \cmidrule(lr){5-8} \cmidrule(lr){9-10} \cmidrule(lr){11-11} \cmidrule(lr){12-13} \cmidrule(lr){14-15}
				& \textit{NYC} & \textit{CHI} & \textit{DC} & \textit{XA-gr.}  & \textit{CD-gr.} & \textit{XA-di.} & \textit{CD-di.} & \textit{SH} & \textit{CQ} & \textit{BJ} & \textit{LA} & \textit{Bay} & \textit{AvgNRMSE} &\textit{WstNRMSE}  \\
				\midrule
				\textbf{Temporal} &&&&&&&&&&&&&&\\
				\emph{HM (TC)}           & 3.206 &	2.458 &	2.304 &	5.280 &	6.969  & 19.893 &	32.098  &  269.16 &	221.39 &	0.768 &	9.471 &	4.155  & 1.865 &	1.909  \\
				\emph{ARIMA (TC)}           & 3.178 & 2.428 & 2.228 & 5.035 & 6.618 & 19.253 &	26.131 &  212.01 & 180.53 & 0.755 & 9.230 &3.936 & 1.667 & 3.687  \\
				\emph{LSTM  (TC) }     		& 3.018 & 2.493 & 2.212 & 4.950 & 6.444 & 18.150 &	23.075& 195.60 & 104.61 & 0.755 & 7.866 &3.683 & 1.463 & 2.596  \\
				\emph{HM (TM)}       		& 2.686 & 2.230 & 1.956 & 4.239 & 4.851  & 16.281 &	17.264 & 108.59 & 74.55 & 0.864 & 9.560 &3.965 & 1.235 & 1.523  \\
				\emph{XGBoost (TM)}  		& 2.704 & 2.376 & 1.956 & 4.172 & 4.915  & 15.040 &	16.766 &81.82 & 69.50 & 0.686 & 8.298 &3.253 & 1.134& 1.420  \\
				\emph{GBRT (TM)}	   		& 2.682 & 2.355 & 1.928 & 4.135 & 4.873  & 16.202 &	\textbf{14.924*} & 83.94 & 72.99 & 0.689 & 8.269 &3.370 & 1.139 & 1.491  \\
				\emph{TMeta-LSTM-GAL (TM)}     		& 2.511 & \textbf{2.133*} & 1.927 & 3.847 & 4.678  & \textbf{12.687} &	15.324 &85.19 & 53.18  & 0.686 & 7.436 &3.231 & 1.047 & 1.130 \\
				\midrule
				\textbf{Temporal \& Spatial} &&&&&&&&&&&& \\
				\emph{DCRNN (TC+SP)} 		& 2.618 & 2.246 & 2.118 & 4.529 & 6.258 & 19.487 &	22.945 & 116.15 & 65.72 & 0.757& 8.562 &6.198  & 1.350 &	2.051  \\
				\emph{STGCN (TC+SP)} 		& 2.841 &	2.482 &	2.067 &	3.992 &	5.051  & 14.139 	& 17.777  & 91.29 &	58.34 &	0.694 &	7.871 &	3.136   & 1.130 &	1.211  \\
				\emph{GMAN (TC+SP)} 		& 2.792 &	2.336 &	\textbf{1.836*} &	4.026 &	5.293   & 13.994 	& 20.157  & 97.58 &	51.37 	& 0.764 &	7.276& 	3.688    & 1.142 &	1.351  \\
				\emph{Graph-WaveNet (TC+SP+SD)} 		& 2.666 &	2.158 &	1.874 &	3.986 &	5.097   & 13.682 &	17.170  & 92.88 &	52.52 &	0.719 &	\textbf{6.809*}& 	3.589    & 1.092 &	1.232  \\
				\emph{ST-ResNet (TM+SP)}   	&  ---  &  ---  &  ---  & 3.903 & 4.673  &  ---   &  --- & --- &---  &  --- & --- & ---  & --- & ---\\
				%\emph{ASTGCN (TM+SP)} \\
				\emph{ST-MGCN (TM+SM)}& 2.513 & 2.177 & 1.903 & 3.886 & 4.732  &13.107 &	15.404 & 88.76 & 50.96 & 0.691 & 8.079 &\textbf{3.042} & 1.056 &	1.186 \\
				\emph{AGCRN-CDW (TM+SD)} 		& 2.830 &	2.565 &	2.074 &	3.958 &	4.753    & 16.921 &	17.982  & 238.99 &	131.55 &	0.688 &	8.575& 	\textbf{3.022*}     & 1.440 	& 3.171  \\
				\emph{STMeta-GCL-GAL (TM+SM)}		& \textbf{2.410*} & 2.170 & 1.856 & \textbf{3.808} & 4.650  & \textbf{12.679*} 	&15.307 & \textbf{75.36*} & \textbf{49.47}  & \textbf{0.670}& 7.156 &3.116  & \textbf{1.014*} 	& \textbf{1.051*}	 \\
				\emph{STMeta-GCL-CON (TM+SM)} 	& \textbf{2.411} & \textbf{2.133*} & 1.859 & \textbf{3.772*} & \textbf{4.613*}  & 12.737 &	\textbf{15.227} & 80.69  & 50.01  & \textbf{0.667*}& \textbf{6.889*} &3.204  & \textbf{1.017} &	1.071 \\
				\emph{STMeta-DCG-GAL (TM+SM)}  & \textbf{2.411} & 2.182 & \textbf{1.852} & 3.833 & \textbf{4.635}  & 12.703 &	15.398 & \textbf{77.49}  & \textbf{48.96*} & \textbf{0.670}& 7.184 &3.187  & 1.019 	& \textbf{1.055} \\
				\bottomrule
		\end{tabular}}
	\end{center}
\end{table*}

\begin{table*}[h!]
	\caption{15-minute prediction error. The best two results are highlighted in bold, and the top one result is marked with `*'. EV dataset is collected in a 30-minute frequency, so we do not have its results for 15-minute. (TC: Temporal Closeness; TM: Multi-Temporal Factors; SP: Spatial Proximity; SM: Multi-Spatial Factors; SD: Data-driven Spatial Knowledge Extraction)}
	\label{tab:result-all-15}
	\begin{center}
		
		\resizebox{1\textwidth}{!}{
			\begin{tabular}{lccccccccccccc}
				\toprule
				& \multicolumn{3}{c}{\textbf{Bikesharing}} & \multicolumn{4}{c}{\textbf{Ridesharing}} & \multicolumn{2}{c}{\textbf{Metro}} & \multicolumn{2}{c}{\textbf{Speed}} & \multicolumn{2}{c}{\textbf{Overall}} \\
				\cmidrule(lr){2-4} \cmidrule(lr){5-8} \cmidrule(lr){9-10} \cmidrule(lr){11-12} \cmidrule(lr){13-14}
				& \textit{NYC} & \textit{CHI} & \textit{DC} & \textit{XA-gr.}  & \textit{CD-gr.} & \textit{XA-di.} & \textit{CD-di.} & \textit{SH} & \textit{CQ}  & \textit{LA} & \textit{Bay} & \textit{AvgNRMSE} &\textit{WstNRMSE}  \\
				\midrule
				\textbf{Temporal} &&&&&&&&&&&&&\\
				\emph{HM (TC)}           & 1.903 &	1.756 &	1.655 &	3.155 &	4.050  & 10.022 &	13.530  &93.81 &	76.67 	&	7.150 &	2.967  & 1.461 &	2.443 \\
				\emph{ARIMA (TC)}           & 1.874 & 1.784 & 1.689 & 3.088 & 3.948 & 9.664 &	13.138 &83.54 & 67.11 & 7.028 & 2.869 & 1.394 & 2.138 \\
				\emph{LSTM  (TC) }     		& 1.989 & 1.802 & 1.678 & 3.051 & 3.888 & 9.640 &	12.367 & 80.40 & 55.37 & 6.380 & 2.690 & 1.332 	& 1.964 \\
				\emph{HM (TM)}       		& 1.892 & 1.668 & 1.555 & 2.828 & 3.347  & 8.883& 	10.650 & 49.75 & 45.26 & 8.934 & 3.690 & 1.246 	& 1.695 \\
				\emph{XGBoost (TM)}  		& 1.712 & 1.672 & 1.559 & 2.799 & 3.430  & 8.444 &	10.368 & 47.89 & 35.70 & 6.443 & 2.623 & 1.115 	& 1.223 \\
				\emph{GBRT (TM)}	   		& 1.708 & 1.667 & 1.552 & 2.775 & 3.363  & 8.511 &	10.310 & 44.55 & 33.29 & 6.371 & 2.645 & 1.097 	& 1.209 \\
				\emph{TMeta-LSTM-GAL (TM)}     		& 1.818 & 1.623 & 1.540 & 2.917 & 3.286  & 7.932 &	9.815 & 45.88 & 33.34  & 6.156 & 2.544 & 1.086 &	\textbf{1.168} \\
				\midrule
				\textbf{Temporal \& Spatial} &&&&&&&&&&&&& \\
				\emph{DCRNN (TC+SP)} 		& 1.712 & 1.718 & 1.594 & 2.889 & 3.743 & 9.584 &	12.197 & 56.00 & 37.07 & 6.440 & 5.322 & 1.285 	& 2.194 \\
				\emph{STGCN (TC+SP)} 		& 1.738 &	1.806 &	1.630 &	2.789 &	3.453  & 8.402 &	11.070  & 47.40 &	35.19 & 6.236 &	2.493  & 1.125 &	1.237 \\
				\emph{GMAN (TC+SP)} &\textbf{1.632*}& 	\textbf{1.529}& 	\textbf{1.355*}& 	2.769& 	3.520  & 8.503 &	11.598 &49.21 &	36.66	& 6.214 &	3.484  & 1.136 &	1.436 \\
				\emph{Graph-WaveNet (TC+SP+SD)} 		& \textbf{1.644} &	\textbf{1.460*} &	\textbf{1.357} &	2.764 &	3.442   & 8.066 &	10.571  & 47.84 &	35.04 	&	\textbf{5.270*} &	2.780   & 1.065 	& 1.169 \\
				\emph{ST-ResNet (TM+SP)}   	&  ---  &  ---  &  ---  & 2.686 & 3.314  &  --- & --- & ---   &  ---  & --- & --- & --- & ---  \\
				%\emph{ASTGCN (TM+SP)} \\
				\emph{ST-MGCN (TM+SM)}& 1.687 & 1.646 & 1.545 & 2.714 & 3.293  & 7.986 &	9.818 & 46.54 & \textbf{32.72} & 6.645 & \textbf{2.426*} & 1.078 &	1.261 \\
				\emph{AGCRN-CDW (TM+SD)} 		& 1.836 &	1.883 &	1.745& 	2.722& 	3.296    & 9.386 &	10.127  & 77.06 &	46.95 	&	6.709 &	2.453    & 1.246 &	1.882 \\
				\emph{STMeta-GCL-GAL (TM+SM)}		& 1.659 & 1.607 & 1.527 & 2.653 & \textbf{3.244}  & \textbf{7.561*} &	\textbf{9.695} & \textbf{41.67} & \textbf{31.39*}  & \textbf{5.644} & \textbf{2.433} & \textbf{1.031*} 	& \textbf{1.127*} \\
				\emph{STMeta-GCL-CON (TM+SM)} 	& 1.673 & 1.629 & 1.512 & \textbf{2.637*} & \textbf{3.241*} & 7.791 &	\textbf{9.673*} & 43.83  & 38.21 & 5.800 & 2.449 & 1.062 	& 1.217 \\
				\emph{STMeta-DCG-GAL (TM+SM)}  & 1.654 & 1.609 & 1.517 & \textbf{2.648} & 3.254  & \textbf{7.717} &	9.716 & \textbf{40.94*} & 36.90 & 5.788 & 2.446 & \textbf{1.050} &	1.176 \\
				\bottomrule
		\end{tabular}}
	\end{center}
\end{table*}

\newpage
\subsection*{Dataset Details}

\begin{figure*}[b]%[tbhp]
	\centering
	\begin{subfigure}[t]{.19\linewidth}
		\includegraphics[width=1\linewidth]{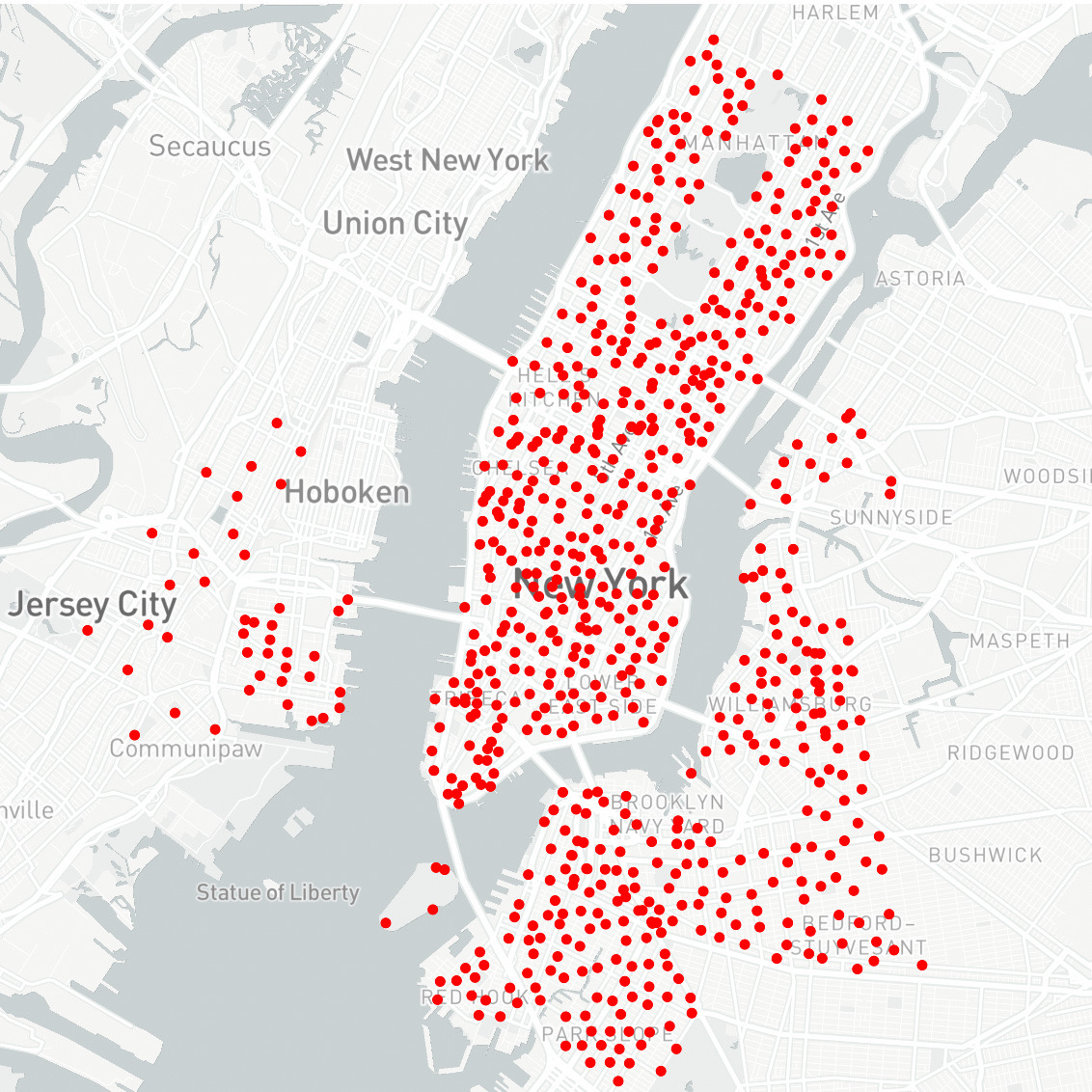}
		\caption{BS: NYC}
		\label{fig:nyc}
	\end{subfigure}
	\begin{subfigure}[t]{.19\linewidth}
		\includegraphics[width=1\linewidth]{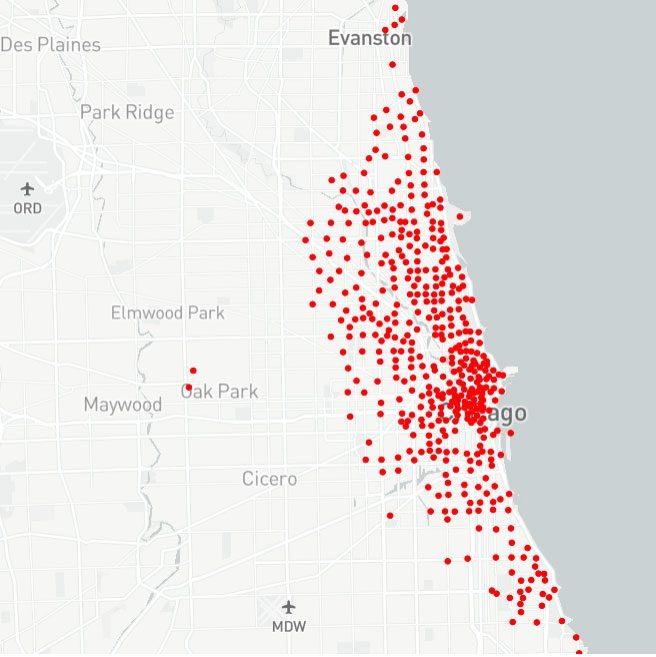}
		\caption{BS: CHI}
		\label{fig:chicago}
	\end{subfigure}
	\begin{subfigure}[t]{.19\linewidth}
		\includegraphics[width=1\linewidth]{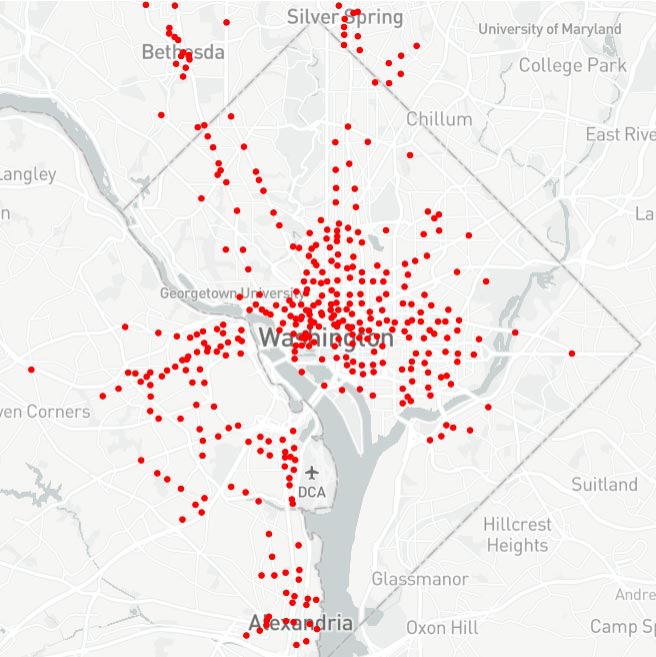}
		\caption{BS: DC}
		\label{fig:dc}
	\end{subfigure}
	\begin{subfigure}[t]{.19\linewidth}
		\includegraphics[width=1\linewidth]{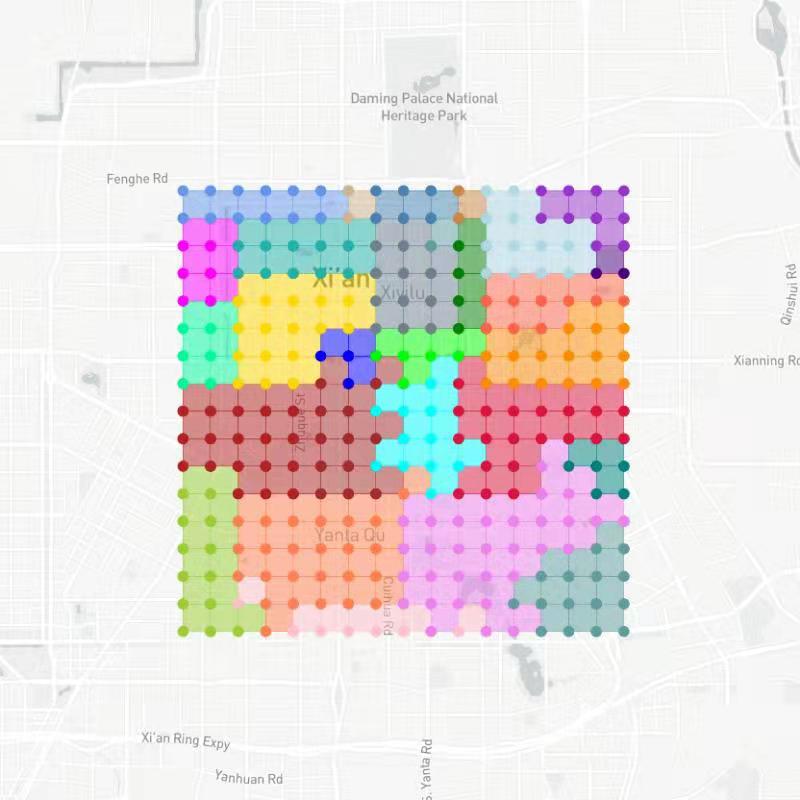}
		\caption{RS: XA (gr. \& di.)}
		\label{fig:xian}
	\end{subfigure}
	\begin{subfigure}[t]{.19\linewidth}
		\includegraphics[width=1\linewidth]{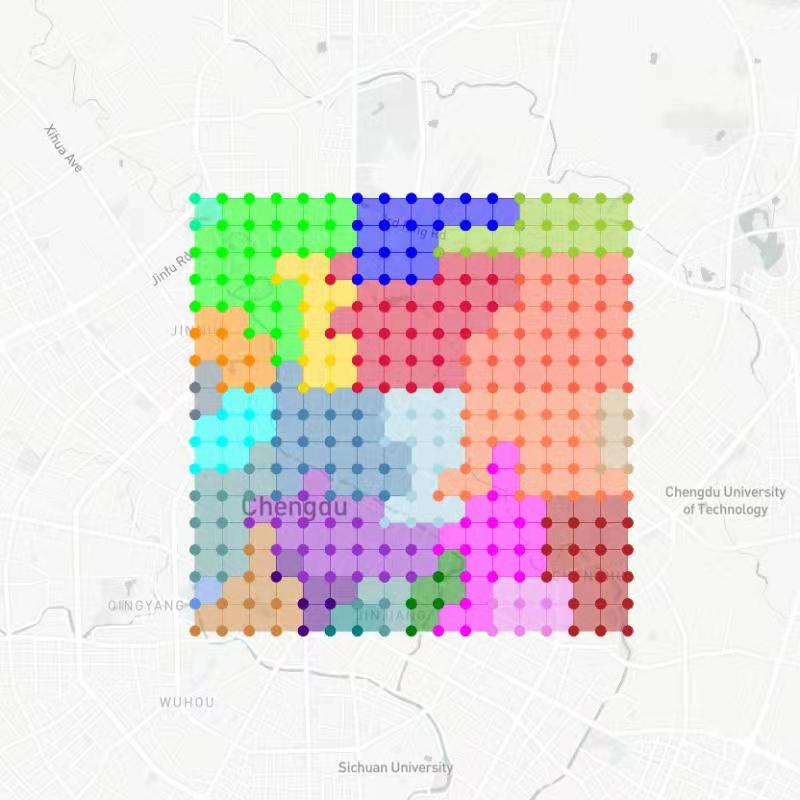}
		\caption{RS: CD (gr. \& di.)}
		\label{fig:chengdu}
	\end{subfigure}
	\\
	\begin{subfigure}[t]{.19\linewidth}
		\includegraphics[width=1\linewidth]{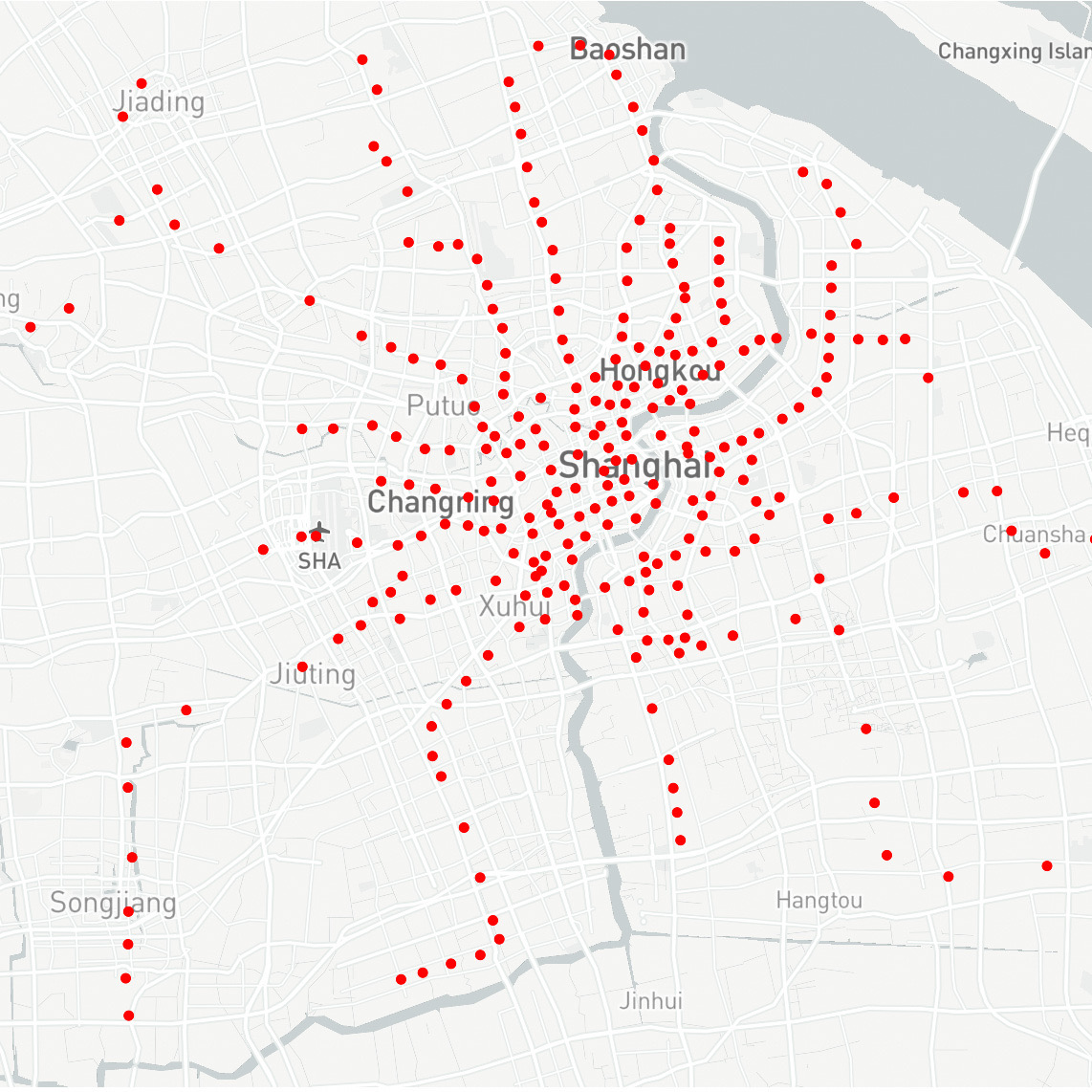}
		\caption{MT: SH}
		\label{fig:shanghai}
	\end{subfigure}
	\begin{subfigure}[t]{.19\linewidth}
		\includegraphics[width=1\linewidth]{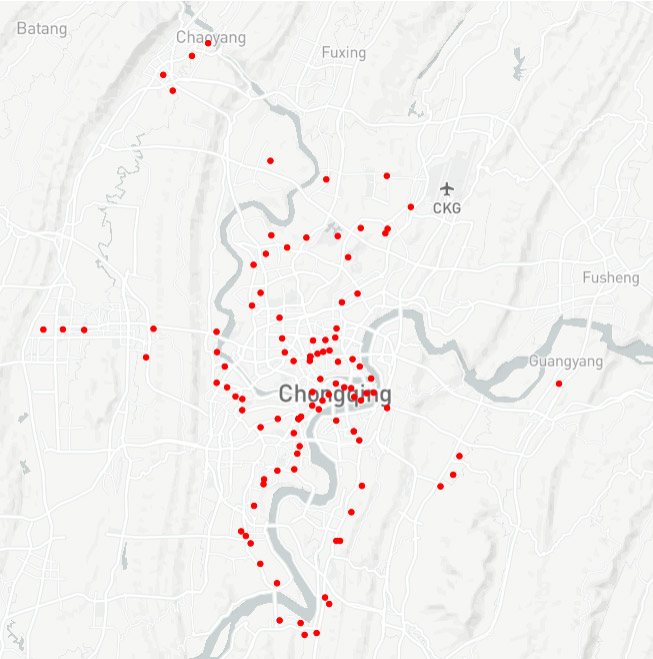}
		\caption{MT: CQ}
		\label{fig:chongqing}
	\end{subfigure}
	\begin{subfigure}[t]{.19\linewidth}
		\includegraphics[width=1\linewidth]{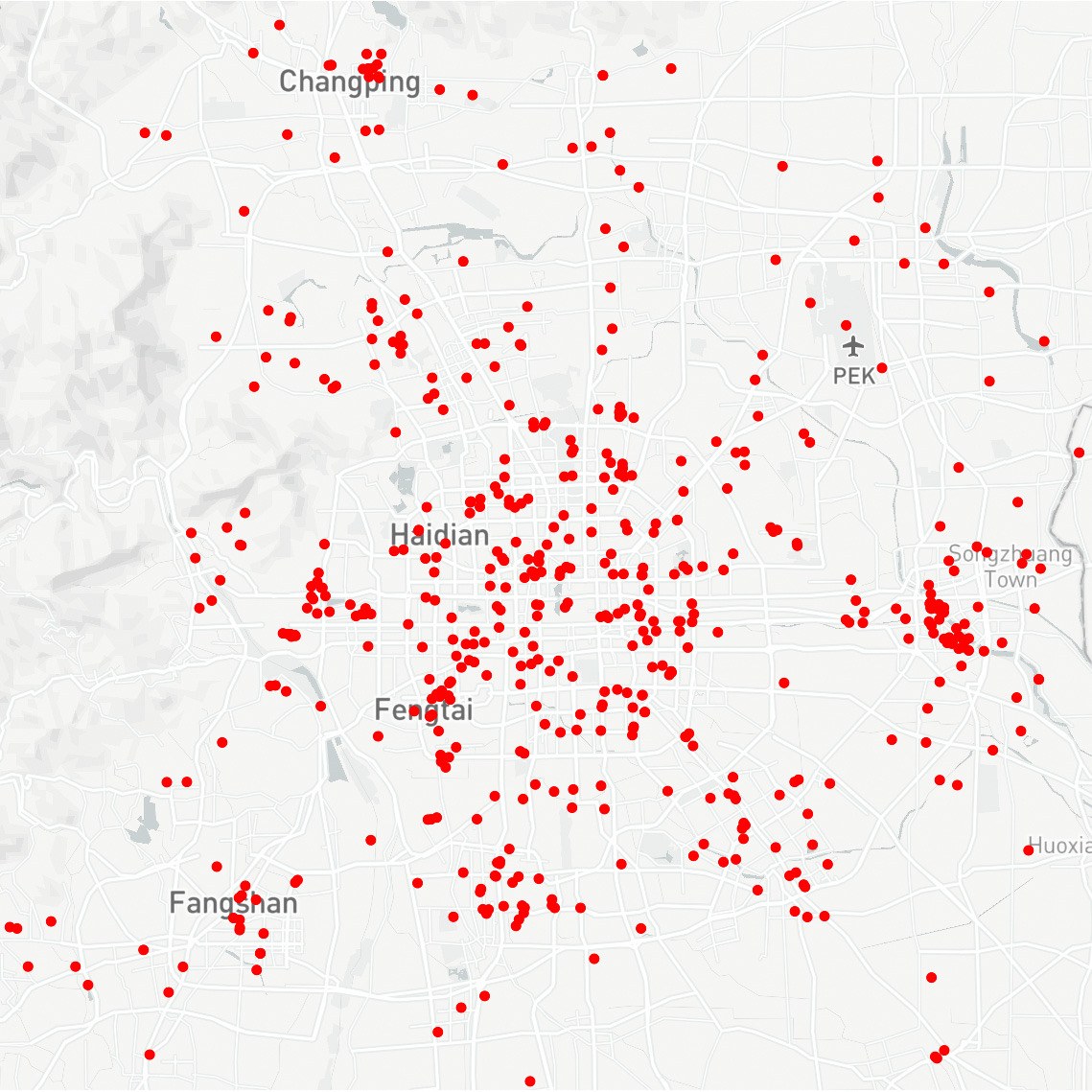}
		\caption{EV: BJ}
		\label{fig:beijing}
	\end{subfigure}
	\begin{subfigure}[t]{.19\linewidth}
		\includegraphics[width=1\linewidth]{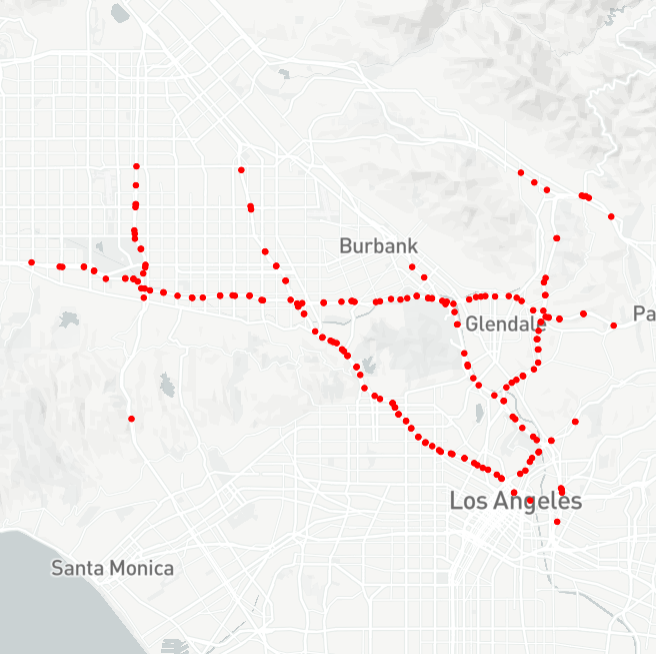}
		\caption{SP: LA}
		\label{fig:la}
	\end{subfigure}
	\begin{subfigure}[t]{.19\linewidth}
		\includegraphics[width=1\linewidth]{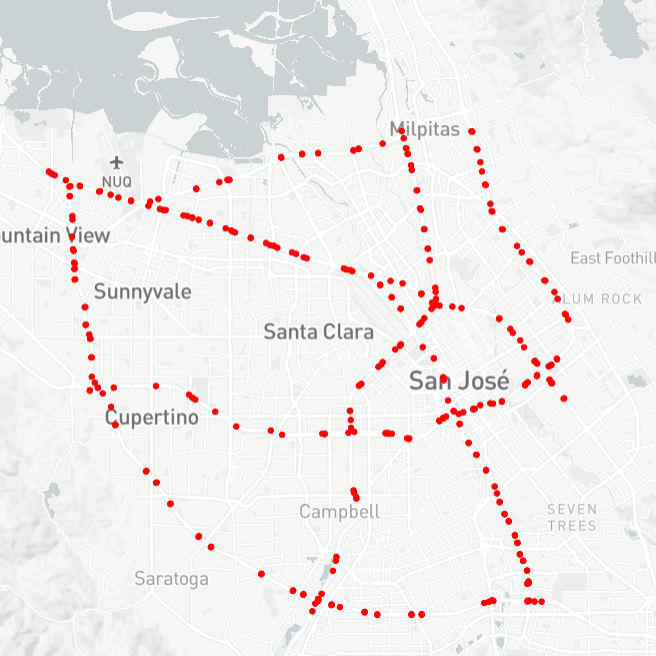}
		\caption{SP: Bay}
		\label{fig:bay}
	\end{subfigure}
	\caption{Visualization of each dataset. (BS: bikesharing, RS: ridesharing, MT: metro, EV: electrical vehicle, SP: speed)}
	\label{fig:cities}
\end{figure*}

The city area and the locations for each dataset are shown in Figure~\ref{fig:cities}. The dataset statistics are listed in Table~\ref{tab:datasets}.

1) \textbf{Bikesharing}. The bikesharing datasets are collected from U.S. open data portals including New York City (NYC, \url{https://www.citibikenyc.com/system-data}), Chicago (CHI, \url{https://www.divvybikes.com/system-data}), and DC (\url{https://www.capitalbikeshare.com/system-data}). The dataset time span for all three cities is more than four years. The total number of historical flow records is around 49 million, 13 million, and 14 million in NYC, Chicago, and DC, respectively, and each record contains the start station, start time, stop station, stop time, etc. We predict the number of bikesharing demands in each station (i.e., the number of bike borrowers). %The dataset statistics are summarized in Table \ref{tab:bike}. %We also collect weather data from NCEI website (National Centers for Environmental Information).

%\subsection{Ridesharing}
2) \textbf{Ridesharing}. The ridesharing order datasets are collected from DiDi's open research collaboration project, including the Chinese cities of Xi'an (XA) and Chengdu (CD). The time span is two months, and the total number of the historical ridesharing orders is around 6 and 8 million for Xi'an and Chengdu, respectively. The order records contain start location, start time, end location, and end time. The location information has longitude and latitude. {blue} Note that these open data cover only the central area of Xi'an and Chengdu, instead of the whole city area. We use two different strategies to split regions for prediction. (1) \textit{Grid (gr.)}: \color{black}
We divide the area into $16*16$ grids with a size of $0.5km*0.5km$ for each grid, then we predict the number of taxi orders in each grid;  (2) \textit{District (di.)}: We split the area according to the administrative district division, and obtain 28 and 25 districts for Xi'an and Chengdu, respectively.

\begin{table*}[b!]
	\footnotesize
	\caption{Dataset statistics}
	\label{tab:datasets}
	\begin{center}
		\begin{tabular}{lccccc}
			\toprule
			& \multicolumn{3}{c}{\textbf{Bikesharing}} & \multicolumn{2}{c}{\textbf{Ridesharing}} \\
			\cmidrule(lr){2-4} \cmidrule(lr){5-6} 
			& \textit{NYC} & \textit{CHI} & \textit{DC} & \textit{XA} & \textit{CD} \\
			\midrule
			\textbf{Time span}     & 2013.07-2017.09 & 2013.07-2017.09 & 2013.07-2017.09 & 2016.10-2016.11 & 2016.10-2016.11 \\
			\textbf{\#Records}		& 49,100,694 & 13,130,969 & 13,763,675 & 5,922,961 & 8,439,537 \\
			\textbf{\#Locations}     & 820 & 585 & 532 & 256 (grid), 28 (district)  & 256 (grid), 25 (district) \\
			\bottomrule
		\end{tabular}
		\\ \vspace{+2em}
		\begin{tabular}{lccccc}
			\toprule
			& \multicolumn{2}{c}{\textbf{Metro}} & \textbf{EV} & \multicolumn{2}{c}{\textbf{Speed}} \\
			\cmidrule(lr){2-3} \cmidrule(lr){4-4} \cmidrule(lr){5-6}
			& \textit{Shanghai} & \textit{Chongqing} & \textit{Beijing} & \textit{LA} & \textit{Bay}\\
			\midrule
			\textbf{Time span} & 2016.07-2016.09 & 2016.08-2017.07 & 2018.03-2018.08 & 2012.03-2012.06 & 2017.01-2017.07\\
			\textbf{\#Records} & 333,149,034 & 409,277,117 & 1,272,961 & 34,272 & 52,128\\
			\textbf{\#Locations} & 288 & 113 & 629 & 207 & 325\\
			%\#FrequentStations     & 717 & 444 & 378\\
			\bottomrule
		\end{tabular}
	\end{center}
\end{table*}

%The dataset statistics are summarized in Table \ref{tab:didi}.

%\begin{table}[t]
%	\small
%	\caption{Ridesharing dataset statistics}
%	\label{tab:didi}
%	\begin{minipage}{\columnwidth}
	%		\begin{center}
		%			\begin{tabular}{lll}
			%				\toprule
			%				& Xi'an & Chengdu \\
			%				\midrule
			%				Time span     & 2016.10.01-2016.11.30 & 2016.10.01-2016.11.30 \\
			%				\#Orders & 5,922,961 & 8,439,537 \\
			%				\#Grids     & 256 & 256 \\
			%				%\#FrequentGrids     & 253 & 256 \\
			%				\bottomrule
			%			\end{tabular}
		%		\end{center}
	%	\end{minipage}
%\end{table}

%\subsection{Metro}

3) \textbf{Metro}. The metro datasets are collected from Shanghai (SH) and Chongqing (CQ). For Shanghai, the time span is three months with around 333 million records; for Chongqing, the time span is one year with around 409 million records. Each metro trip record has the check-in time, check-in station, check-out time, and check-out station. We target predicting the check-in flow amount for all the metro stations. %The dataset statistics are summarized in Table \ref{tab:metro}.

%\begin{table}[t]
%	\small
%	\caption{Metro dataset statistics}
%	\label{tab:metro}
%	\begin{minipage}{\columnwidth}
	%		\begin{center}
		%			\begin{tabular}{lll}
			%				\toprule
			%				& Shanghai & Chongqing \\
			%				\midrule
			%				Time span     & 2016.07.01-2016.09.14 & 2016.08.01-2017.07.31 \\
			%				\#Check-in Records & 333,149,034 & 409,277,117 \\
			%				\#Stations     & 288 & 113 \\
			%				%\#FrequentStations & 288 & 113 \\
			%				\bottomrule
			%			\end{tabular}
		%		\end{center}
	%	\end{minipage}
%\end{table}

4) \textbf{Electrical Vehicle (EV)}. The EV charging station usage dataset is collected from Beijing (BJ) in the form of stations' occupation at different time slots, i.e., the number of available and occupied docks. The holder of the stations is one of the largest EV charging station companies in China. The dataset time span is six months and the total number of EV charging station usage records is more than one million.  We predict the number of docks in use for each station as this is a demand indicator of the EV charging stations. %The dataset statistics are summarized in Table \ref{tab:cs}.

5) \textbf{Speed}. The two traffic speed datasets are widely used in STTP research: METR-LA and PEMS-BAY from Los Angeles (LA) County and Bay Area, respectively. In METR-LA, 207 sensors record highway vehicles' speeds for four months; In PEMS-BAY, there are 325 sensors for six months. Each sensor can be seen as a station, and we predict the traffic speed of each sensor at the next time slot.

\end{document}